\documentclass[10pt,twocolumn,letterpaper]{article}

\usepackage{cvpr}              
\usepackage{caption}
\usepackage{graphicx}   
\usepackage{microtype}      
\usepackage{xcolor}         
\usepackage[table]{xcolor}
\usepackage{enumitem}
\usepackage{colortbl}
\usepackage{multirow}
\usepackage{tcolorbox}
\usepackage{wrapfig}
\newtcolorbox{promptbox}[1][]{
    colback=gray!10!white,
    colbacktitle=white,
    coltitle=black,
    colframe=black!75!black,
    boxrule=0.7pt,
    halign title=center,
    title=\textbf{#1}
}

\definecolor{cvprblue}{rgb}{0.21,0.49,0.74}
\usepackage[pagebackref,breaklinks,colorlinks,allcolors=cvprblue]{hyperref}

\def\paperID{***} 
\def\confName{CVPR}
\def\confYear{2026}

\begin{document}

\newcommand\blfootnote[1]{%
  \begingroup
  \renewcommand\thefootnote{}\footnote{#1}%
  \addtocounter{footnote}{-1}%
  \endgroup
}

\title{Unified Multimodal Models as Auto-Encoders}

\author{
 {Zhiyuan Yan\textsuperscript{1,2}$^{\diamond,\star}$},
 {Kaiqing Lin\textsuperscript{1}$^{\diamond}$},
 {Zongjian Li\textsuperscript{1,3}$^{\diamond}$},
 {Junyan Ye\textsuperscript{4}$^{\diamond}$},
 {Hui Han\textsuperscript{1}},
 {Haochen Wang\textsuperscript{2,6}$^{\star}$},
\\
 {Zhendong Wang\textsuperscript{5}},
 {Bin Lin\textsuperscript{1,3}},
 {Hao Li\textsuperscript{1}},
 {Xinyan Xiao\textsuperscript{2}},
 {Jingdong Wang\textsuperscript{2}},
 {Haifeng Wang\textsuperscript{2}},
 {Li Yuan\textsuperscript{1}$^{\dagger}$}
\\
\\
 \textsuperscript{1}Shenzhen Graduate School, Peking University \\
 \textsuperscript{2}Baidu,
 \textsuperscript{3}Rabbitpre AI,
 \textsuperscript{4}SYSU,
 \textsuperscript{5}USTC,
 \textsuperscript{6}CASIA
\\
 \small\tt{
   \href{mailto:zhiyuanyan@stu.pku.edu.cn}{zhiyuanyan@stu.pku.edu.cn}
 }
}

\twocolumn[{
\renewcommand\twocolumn[1][]{#1}
\maketitle

\begin{center}
\centering
\includegraphics[width=0.95\linewidth]{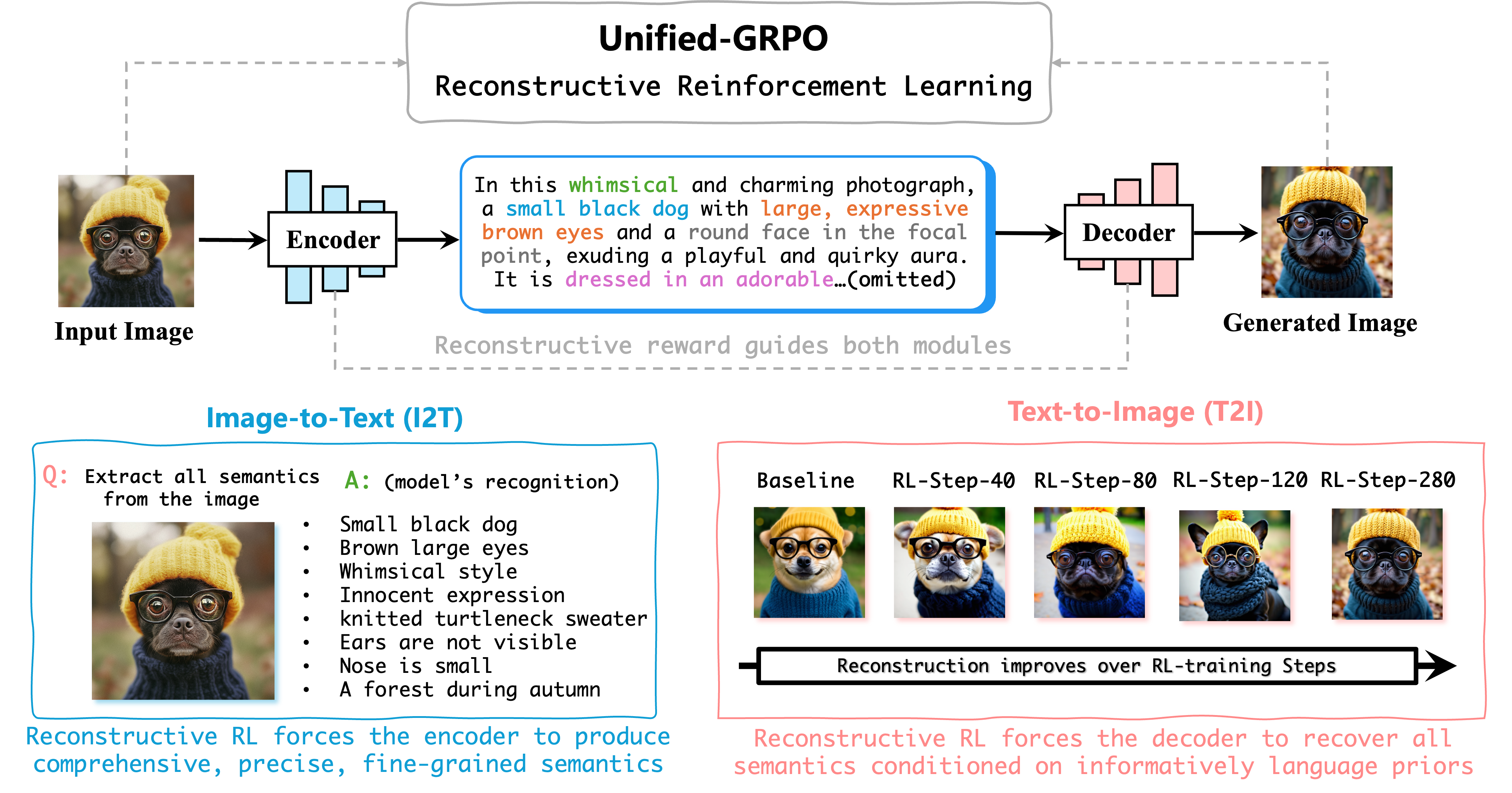}
\captionof{figure}{
Illustration of the key insight of our \textbf{\texttt{UAE}}, unifies image-to-text understanding and text-to-image generation under a reconstructive auto-encoding perspective. 
By optimizing reconstruction similarity through RL, the encoder (und. module) is trained to learn richer semantic representations while the decoder (gen. module) becomes better at recovering all semantics. 
Illustrations show strengthened fine-grained visual perception and improved complex instruction-following generation capability across RL training steps.
}
\small
\label{fig:insight}
\end{center}
}]
\maketitle

\begin{abstract}

\blfootnote{
$\diamond$ Equal Contribution,
$\star$ Work done during an internship at the Baidu Star Program,
$^\dagger$ Corresponding Author
}

Image-to-text (I2T) understanding and text-to-image (T2I) generation are two fundamental, important yet traditionally isolated multimodal tasks. 
Despite their intrinsic connection, existing approaches typically optimize them independently, missing the opportunity for mutual enhancement.
In this paper, we argue that the both tasks can be connected under a shared Auto-Encoder perspective, where text serves as the intermediate latent representation bridging the two directions — encoding images into textual semantics (I2T) and decoding text back into images (T2I).
Our key insight is that \textit{if the encoder truly ``understands" the image, it should capture all essential structure, and if the decoder truly ``understands" the text, it should recover that structure faithfully.}
Building upon this principle, we propose Unified-GRPO, a post-training method based on reinforcement learning that jointly optimizes both modules through reconstructive rewards, maximizing the semantic consistency between the input and the generated images.
Under this reconstruction objective, the encoder is encouraged to extract as much accurate and comprehensive semantic information from the input image to maximize reconstruction quality, while the decoder is simultaneously optimized to generate conditioned on the encoder's prior, enabling a self-evolving improvement.

Empirically, we find that using text as the intermediate representation and training under a reconstructive RL paradigm effectively benefits both I2T and T2I.
The I2T module gains stronger fine-grained visual perception, such as small-object recognition, grounding, etc, while its dense embeddings and language priors, in turn, provide richer semantic signals that improve T2I fidelity and complex instruction following.
These results demonstrate that the reconstructive RL establishes a mutually reinforcing cross-modal synergy within the auto-encoding framework.
\end{abstract}    

\vspace{-3mm}

\section{Introduction and Motivation}
\label{sec:intro}


Unified multimodal models (UMMs) that support both generation and understanding have recently gained increasing popularity in both academia and industry~\citep{wang2024emu3, chen2025janus, wu2025janus, xieShowoOneSingle2024, pan2025metaquery, gupta2022metamorph, zhou2024transfusion, yan2025gpt}. 
However, directly combining the understanding and generation models together leads to a sub-optimal result, as most existing arts on UMMs~\citep{wu2025janus, pan2025metaquery, chen2025blip3} suggest that optimizing diffusion-based generative objectives negatively degrade the understanding capability and learned representations (and conversely), making joint training brittle.

Consequently, some existing works decouple the UMM problem~\citep{wu2025janus,qu2025tokenflow}, training understanding and generation modules separately, and missing out on potential cross-task mutual benefits.
These design choices and empirical observations have dampened confidence in truly unified systems: absent demonstrable mutual gains, ``unification" collapses into training two large components side by side. 


In this work, we revisit the relationship between I2T and T2I from a conceptual standpoint and argue that a more principled linkage can be established by viewing them through a unified Auto-Encoder (AE) perspective.
Under this view, text acts as an intermediate latent representation: the encoder extracts a semantic description from the input image (I2T), and the decoder reconstructs an image from this semantic representation (T2I). This perspective offers a natural and powerful unifying principle: \textbf{if the encoder genuinely understands the image, it should capture all essential visual structure; if the decoder genuinely understands the text, it should faithfully recover that structure.}
Thus, high-quality reconstruction becomes a proxy for enhancing both tasks simultaneously, revealing a pathway toward bidirectional synergy.

Building upon this insight, we introduce \textbf{Unified-GRPO}, a reinforcement-learning-based post-training method that jointly optimizes the encoder and decoder through reconstructive rewards.
Unified-GRPO maximizes the semantic consistency between the original and reconstructed images, encouraging the encoder to produce richer and more accurate textual semantics, while guiding the decoder to generate images that better adhere to the encoder’s descriptions.
Through this cross-modal feedback loop, the two modules co-evolve: \textbf{the encoder learns to encode more informative and comprehensive representations, and the decoder learns to generate more faithful and semantically grounded images, creating a self-reinforcing improvement cycle.}

\begin{figure}[!t]
    \centering
    \includegraphics[width=\linewidth]{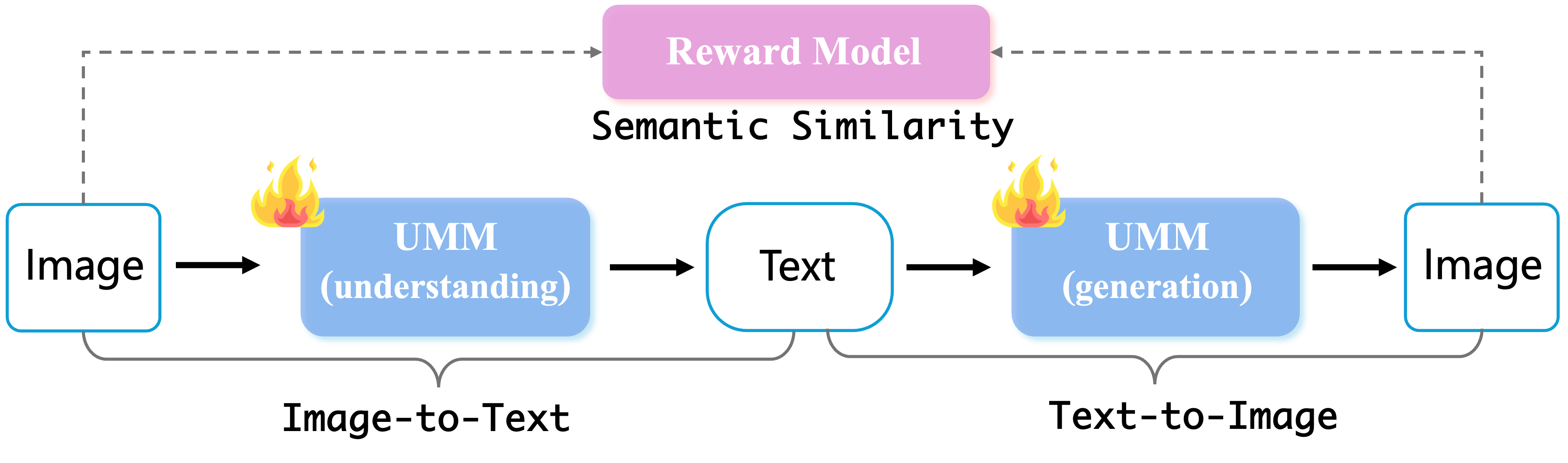}
    \caption{\textbf{The overall workflow of our method.} Our post-training method, Unified-GRPO, utilizes the reconstruction objective for improved unified multimodal models (UMMs). 
    }
    \vspace{-1mm}
    \label{fig:three_stage}
    \vspace{-3mm}
\end{figure}

We conduct extensive experiments on visual understanding, generation, and unification tasks across a broad suite of benchmarks to verify that our post-training strategy with our core AE insight can improve the UMMs~\cite{lin2025uniworld,chen2025janus}.
Specifically, our method achieves significant improvement on image generation (\textit{e.g.}, from 0.73→0.86 on GenEval~\citep{ghosh2023geneval} and 0.296→0.475 on GenEval++~\citep{ye2025echo}), and largely improved fine-grained visual recognition and perception capability, \textit{e.g.}, from 0.05→0.45 on small object detection and from 0.15→0.75 on Person ReID of the MMT-Bench~\citep{mmt}, consistent with the findings reported by Ross~\citep{wang2024reconstructive}) while maintaining the overall performance across visual understanding tasks. Furthermore, results on the proposed Unified-Bench show that our post-training method can largely improve the unification, resulting in a more coherent information flow between encoding and decoding.

In summary, our work makes the following contributions:
\begin{itemize}[nolistsep, leftmargin=*]
    \item \textbf{A unified Auto-Encoder perspective linking I2T and T2I:} We propose a principled formulation where text serves as the intermediate representation connecting image encoding and decoding, offering a coherent bridge between multimodal understanding and generation.

    \item \textbf{Unified-GRPO, an RL-based post-training framework for cross-modal self-evolution:} Through reconstructive rewards, our method jointly optimizes the encoder and decoder, enabling mutual reinforcement: richer semantic encoding improves generation, and more faithful generation strengthens fine-grained visual perception.
    
    \item \textbf{Broad applicability and consistent empirical gains:} Unified-GRPO applies to various encoder–decoder multimodal systems, consistently improving text-to-image generation and enhancing fine-grained understanding (e.g., grounding, small-object recognition), while revealing interpretable trade-offs in OCR-heavy scenarios.
\end{itemize}

\section{\texttt{UAE} Methodology}

Our goal is to unify image-to-text (I2T) understanding and text-to-image (T2I) generation within a single auto-encoding perspective, where text serves as the intermediate latent representation connecting the two directions. Given an input image $x$, a unified multimodal model (UMM) first produces a semantic description $y$ (I2T), and another UMM reconstructs an image $\hat{x}$ from $y$ (T2I). We then adopt reconstructive reinforcement learning to maximize the semantic similarity between $x$ and $\hat{x}$, enabling mutual improvement between understanding and generation.

\begin{figure}[!t]
    \centering
    \includegraphics[width=\linewidth]{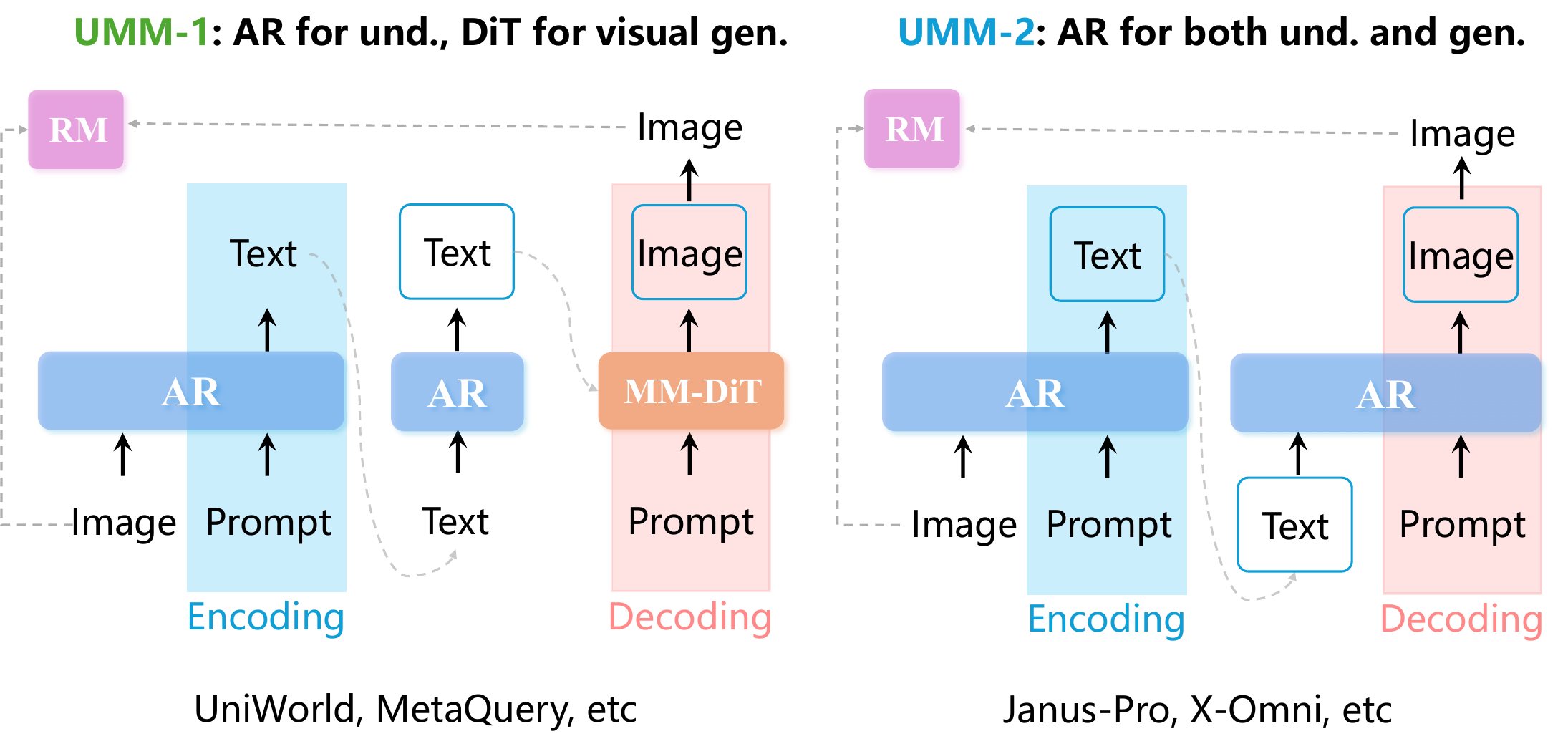}
    \caption{
    \textbf{Illustration of how Unified-GRPO integrates into two representative UMM architectures.} 
    UMM-1 uses an AR model for understanding and an MM-DiT for image generation, while UMM-2 employs a single AR backbone for both understanding and generation. 
    In the diagram, ``text’’ or ``image’’ inside a rectangle denotes latent tokens, whereas those without a rectangle
    represent raw data inputs. 
    Unified-GRPO can be applied to the LLM backbone in both
    architectures to provide reconstructive RL.
    }
    \vspace{-1mm}
    \label{fig:uae_uni}
    \vspace{-2mm}
\end{figure}

\subsection{Unified-GRPO}

We propose \textbf{Unified-GRPO}, a reconstructive reinforcement learning method designed to
\emph{unify} image-to-text (I2T) understanding and text-to-image (T2I) generation by training the model to maximize reconstruction fidelity.

To apply this framework to existing unified multimodal models (UMMs), we consider the two
dominant architectural families shown in Fig.~\ref{fig:uae_uni}:
(1)~\textbf{UMM-1}, where an autoregressive language model (LLM) is responsible for
multimodal understanding and provides language priors for a diffusion transformer (MM-DiT)
used in image generation (e.g., UniWorld~\cite{lin2025uniworld}, MetaQuery~\cite{pan2025metaquery}, etc);
and (2)~\textbf{UMM-2}, where a single autoregressive model handles both visual understanding
and visual generation in a shared token space (e.g., Janus-Pro~\cite{chen2025janus}, X-Omni~\cite{geng2025x}, etc).
In both families, the LLM plays a central role—either as the core understanding module
(UMM-1) or as the backbone of both understanding and generation (UMM-2).
Since GRPO and related RL algorithms have proven highly effective for training LLMs, we
extend this idea to UMMs and employ GRPO to optimize the LLM components toward improved
cross-modal reconstruction.

\paragraph{Applying Unified-GRPO to UMM-1.}
For UMM-1, the autoregressive LLM $\pi_\phi$ is trained using reconstructive RL, while the
diffusion transformer $p_\theta$ remains \emph{frozen} and acts as part of the reward
environment (together with a CLIP encoder). Given an input image $x$, we sample a group of
$G$ caption sequences $\{y^{(i)}\}_{i=1}^G$ from the old policy $\pi_{\phi_{\text{old}}}$. 
For each caption $y^{(i)}$, we extract its last hidden state $h_T^{(i)}$ and project it into
a diffusion condition $c^{(i)}=g(h_T^{(i)})$, which is then used to synthesize a reconstructed
image $\tilde{x}^{(i)} \sim p_\theta(\cdot \mid c^{(i)})$.
The LLM is updated via the GRPO objective~\cite{shao2024deepseekmath}, where each trajectory
$o_i$ corresponds to the token sequence of $y^{(i)}$, and the probability ratio is
\[
r_t^{(i)}(\phi)
=
\frac{
    \pi_\phi(y^{(i)}_t \mid x, y^{(i)}_{<t})
}{
    \pi_{\phi_{\text{old}}}(y^{(i)}_t \mid x, y^{(i)}_{<t})
}.
\]
This process encourages the LLM to emit hidden representations that maximize the diffusion's reconstruction quality.

\paragraph{Applying Unified-GRPO to UMM-2.}
For UMM-2, the same autoregressive model performs both I2T and T2I.
Unified-GRPO is applied in an identical manner, except that the decoder $D_\phi$ is now
autoregressive rather than diffusion-based.
The RL pipeline becomes: $x \xrightarrow{\pi_\phi} y,
y \xrightarrow{\pi_\phi} \tilde{x},
$
with reconstruction reward 
$\mathcal{R}(x,\tilde{x}) = \cos(f_{\mathrm{CLIP}}(x), f_{\mathrm{CLIP}}(\tilde{x}))$.
This enables a fully AR model to co-evolve its understanding and generation abilities within
a single shared token space. 
The specific implementation here is similar to previous work that incentivizes AR for improved image generation (such as T2I-R1~\cite{jiang2025t2i} and AR-GRPO~\cite{yuan2025ar}), and the key difference is that we use the reconstruction reward between the input and the generated image embeddings, enabling it to optimize both understanding and generation modules jointly.

\subsection{Unified-Bench: A Benchmark tailored for Evaluating the Unified Models}
\label{sec:unified-understanding-generation}

\noindent\textbf{Motivation.}
As illustrated in Fig.~\ref{fig:insight}, we view \emph{understanding} (I$\to$T) and \emph{generation} (T$\to$I) as a closed loop whose two halves should \textit{mutually enhance} each other. Judging image realism alone or caption fidelity alone cannot reveal whether a system is truly \emph{unified}. We introduce a reconstruction-based similarity, \textit{i.e.}, \textbf{Unified-Score}, to directly test whether the semantics extracted during understanding are sufficient for faithful regeneration, and whether regeneration in turn validates the completeness of the understanding.

\noindent\textbf{\textit{Protocol-1}: Evaluation of the unified score from the reconstruction similarity.}
To quantify the unified score, we start from 100 diverse source images. The prompt, used to allow the model to generate cpation, is detailed in Supplementary. The same model then synthesizes an image from its \emph{own} caption. We compute unified scores between the reconstruction and the source using four widely adopted vision backbones, CLIP~\citep{radford2021learning}, LongCLIP~\citep{zhang2024long}, DINO-v2~\citep{oquab2023dinov2}, and DINO-v3~\citep{simeoni2025dinov3}, and report per-backbone similarities and an overall summary.

\noindent\textbf{\textit{Protocol-2}: Quality Evaluation of the model's \textit{output caption} for reconstruction.}
We further evaluate caption quality through pairwise comparisons against various baselines, using four commercial LLM judges: Claude-4.1, GPT-4o, Grok-4, and o4-mini. The prompting strategy is detailed in Supplementary. For evaluation, we use pairwise winning rate (\%), the percentage of times our model is preferred over baselines as the main metric.

\begin{table}[t]
\centering
\caption{\textbf{Ablation study on the proposed post-training on understanding, generation, and unification benchmarks.} We apply our method to the two typical unified multimodal models and show the clear improvement.}
\vspace{-10pt}
\resizebox{1\linewidth}{!}{
\begin{tabular}{l|cc|cc|c}
    \toprule
     \multirow{2}{*}{\textbf{Model}} & \multicolumn{2}{c|}{\textbf{Understanding}} & \multicolumn{2}{c|}{\textbf{Generation}} & \multicolumn{1}{c}{\textbf{Unification}} \\
      & MMB & MMMU & GenEval & DPGBench  & Unified-Score  \\
\midrule
UniWorld & 83.5 & 58.6 & 84.0 & 81.2 & 79.0 \\
\textit{w/} Ours & 84.8 & 58.2 & 89.0 & 86.4 & 86.1 \\
\rowcolor{blue!5}
\textit{\textit{vs.} Baseline} & \bf {\textcolor[rgb]{0., 0.5, 0.}{+1.3\%}} & \bf {\textcolor[rgb]{0.5, 0.5, 0.5}{-0.4\%}} & \bf {\textcolor[rgb]{0, 0.5, 0}{+5\%}} & \bf {\textcolor[rgb]{0, 0.5, 0}{+5.2\%}} & \bf {\textcolor[rgb]{0., 0.5, 0.}{+7.1\%}} \\
\midrule
Janus-pro & 79.2 & 41.0  & 80.0 & 84.2 & 82.8 \\
\textit{w/} Ours & 80.3 & 41.6  & 84.3 & 88.9 & 89.1 \\
\rowcolor{blue!5}
\textit{\textit{vs.} Baseline} & \bf {\textcolor[rgb]{0, 0.5, 0}{+1.1\%}} & \bf {\textcolor[rgb]{0, 0.5, 0}{+0.6\%}} & \bf {\textcolor[rgb]{0, 0.5, 0}{+4.3\%}} & \bf {\textcolor[rgb]{0, 0.5, 0}{+4.7\%}} & \bf {\textcolor[rgb]{0, 0.5, 0}{+6.3\%}} \\
\bottomrule
\end{tabular}
}
\vspace{-4mm}
\label{tab:ablation}
\end{table}

\begin{table}[t]
    \centering
\setlength{\tabcolsep}{5pt}
\caption{\textbf{Protocol-1 of Unified-Bench}: comparing of unified score of different methods on Unified-Bench, the tailored benchmark for evaluating the unification between understanding and generation models in the UMMs. 
}
\vspace{-10pt}
\resizebox{1\linewidth}{!}{
    \begin{tabular}{lcccccc}
        \toprule
        Method & \multicolumn{1}{c}{CLIP} & \multicolumn{1}{c}{LongCLIP} & \multicolumn{1}{c}{DINO-v2} & \multicolumn{1}{c}{DINO-v3} & \multicolumn{1}{c}{Overall} \\
        \midrule
        
        GPT-4o-Image~\citep{gpt4o} &	\underline{90.42} & \textbf{94.37} & \underline{81.74} & \underline{77.27} & \underline{85.95} \\
        \midrule
        
        BAGEL~\citep{bagel} & 88.97 & 93.35 & 78.55 & 73.05 & 83.48 \\
        BLIP-3o~\citep{chen2025blip3} & 84.84 & 90.24 & 68.31 & 62.86 & 76.56 \\
        Janus-Pro~\citep{chen2025janus} & 88.72 & 93.45 & 78.30 & 70.61 & 82.77 \\
        OmniGen2~\citep{wu2025omnigen2} & 88.36 & 93.11 & 77.70 & 74.07 & 83.31 \\
        Show-o~\citep{xie2024show} & 80.18 & 86.75 & 58.20 & 51.51 & 69.16 \\
        UniWorld-V1~\citep{lin2025uniworld} & 85.49 & 91.53 & 72.12 & 66.83 & 78.99 \\

        \midrule

        \rowcolor{blue!5}\textbf{\texttt{UAE}} & \textbf{90.50} & \underline{94.35} & \textbf{81.98} & \textbf{77.54} & \textbf{86.09} \\
        
        \bottomrule
        \end{tabular}
        }
\label{tab:unified-bench}
\vspace{-4mm}
\end{table}

\begin{table}[t]
    \centering
    \caption{\textbf{Evaluating how “friendly” the output caption is for image generation}. We use the data from Unified-Bench to assess the quality of the captions produced by the understanding model for better text-to-image generation. \textbf{Bold} indicates the best result.}
 \setlength{\tabcolsep}{5pt}
\vspace{-10pt}
\resizebox{\linewidth}{!}{
    \begin{tabular}{lcccccc}
        \toprule
        Method & \multicolumn{1}{c}{CLIP} & \multicolumn{1}{c}{LongCLIP} & \multicolumn{1}{c}{DINO-v2} & \multicolumn{1}{c}{DINO-v3} & \multicolumn{1}{c}{Overall} \\
        \midrule
        Qwen-2.5-VL-3B~\citep{bai2025qwen25vl} & 88.34 & 92.62 & 73.91 & 70.02 & 80.85 \\
        Qwen-2.5-VL-7B~\citep{bai2025qwen25vl} & 88.26 & 92.89 & 76.12 & 70.96 & 81.92 \\
        \midrule
        \rowcolor{blue!5}\textbf{\texttt{UAE}} & \textbf{90.50} & \textbf{94.35} & \textbf{81.98} & \textbf{77.54} & \textbf{86.09} \\
        \bottomrule
        \end{tabular}
        }
\label{tab:captioner_for_gen}
\end{table}

\begin{table}[!t]
    \centering
    \caption{\textbf{Benchmarking results of text-to-image generation capability on GenEval~\citep{ghosh2024geneval} benchmark}. `$\dagger$' refers to the methods using the LLM rewriter. \textbf{Bold} indicates the best result, and \underline{underlined} denotes the second best.}
    \vspace{-10pt}
    \resizebox{\linewidth}{!}{
    \setlength\tabcolsep{4pt}
        \begin{tabular}{lcccccccc}
            \toprule
            Method & \multicolumn{1}{c}{Single} & \multicolumn{1}{c}{Two} & \multicolumn{1}{c}{Counting} & \multicolumn{1}{c}{Colors} & \multicolumn{1}{c}{Position} & \multicolumn{1}{c}{Color} & \multicolumn{1}{c}{Overall} \\
            \midrule
            Janus Pro~\citep{chen2025janus} & 0.99 & 0.89 & 0.59 & \underline{0.90} & \textbf{0.79} & 0.66 & 0.80 \\
            BLIP3-o 8B~\citep{chen2025blip3} & - & - & - & - & - & - & 0.84 \\
            UniWorld-V1~\citep{lin2025uniworld} & 0.99 & 0.93 & 0.79 & 0.89 & 0.49 & 0.70 & 0.80 \\
            UniWorld-V1$^\dagger$~\citep{lin2025uniworld} & 0.98 & 0.93 & 0.81 & 0.89 & 0.74 & 0.71 & 0.84 \\
            OmniGen2~\citep{wu2025omnigen2} & \textbf{1.00} & \underline{0.95} & 0.64 & 0.88 & 0.55 & 0.76 & 0.80 \\
            X-Omni$^\dagger$~\citep{geng2025x} & 0.98 & \underline{0.95} & 0.75 & 0.91 & 0.71 & 0.68 & 0.83 \\
            BAGEL~\citep{bagel} & 0.99 & 0.94 & 0.81 & 0.88 & 0.64 & 0.63 & 0.82 \\
            BAGEL$^\dagger$~\citep{bagel} & 0.98 & \underline{0.95} & \textbf{0.84} & \textbf{0.95} & \underline{0.78} & 0.77 & \underline{0.88} \\
            \midrule
            
            \rowcolor{blue!5}\textbf{\texttt{UAE}} & \textbf{1.00} & 0.89 & \textbf{0.84} & \underline{0.90} & 0.71 & \underline{0.79} & 0.86 \\              \rowcolor{blue!8}\textbf{\texttt{UAE}$^\dagger$} & \textbf{1.00} & \textbf{0.97} & \underline{0.82} & \textbf{0.95} & 0.73 & \textbf{0.84} & \textbf{0.89} \\  
            \bottomrule
        \end{tabular}
    }
    \vspace{-5pt}
    \label{tab:geneval}
\end{table}

\begin{figure*}[!t]
    \centering
    \includegraphics[width=0.95\linewidth]{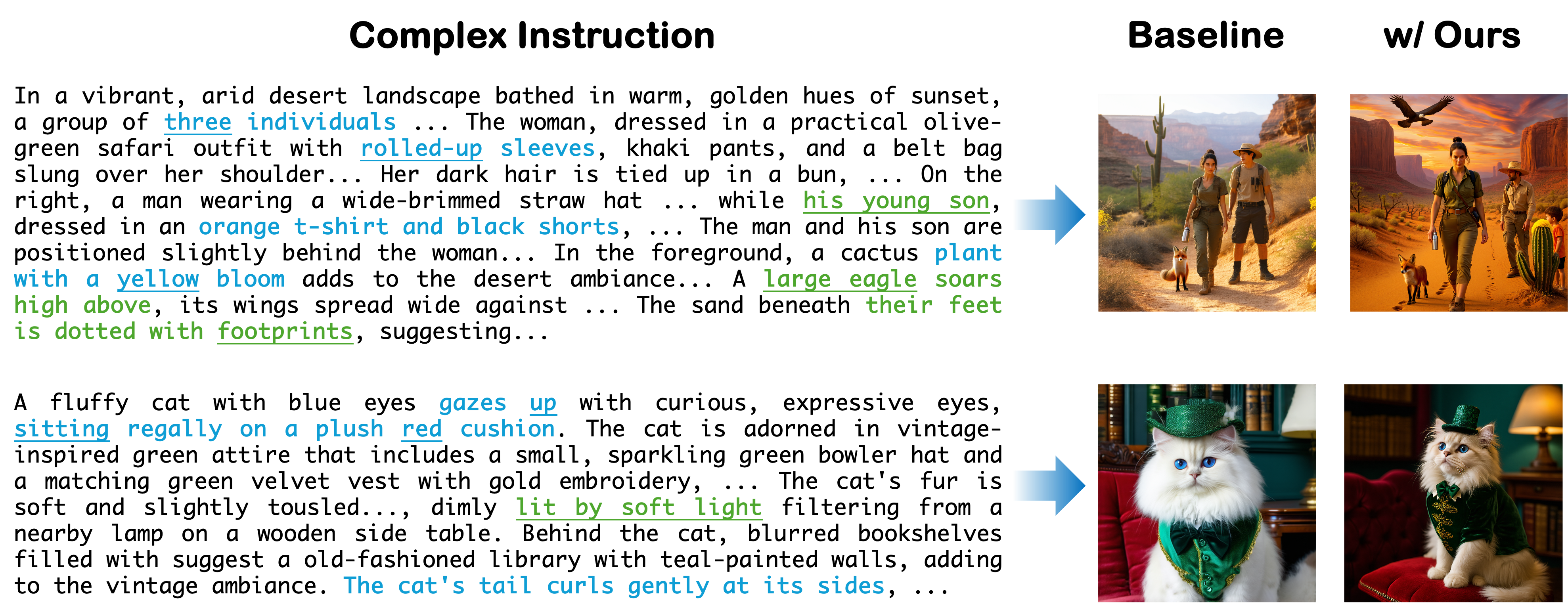}
    \vspace{-10pt}
    \caption{
    \textbf{Qualitative results on the complex and long-context generation.} Our method can recover very detailed semantics from the highly descriptive input caption over the baseline, demonstrating that improved understanding can notably benefit generation.
    }
    \vspace{-4mm}
    \label{fig:complex}
\end{figure*}

\begin{figure*}[!t]
    \centering
    \includegraphics[width=\linewidth]{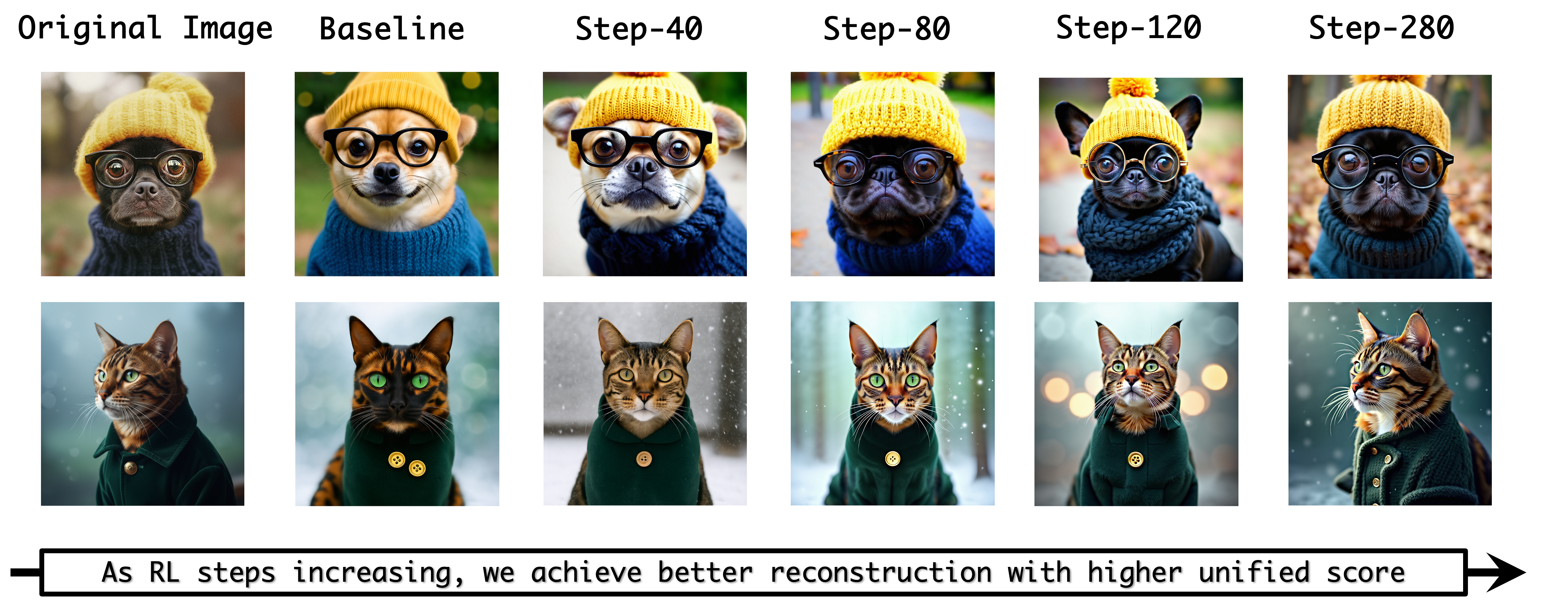}
    \vspace{-15pt}
    \caption{\textbf{Reconstruction results \textit{vs.} RL training steps.} With the RL steps increasing, the understanding model (encoder) achieves better perception capability to produce an informative, detailed, yet accurate description to reconstruct the original image comprehensively; while the generation model (decoder) can take the rich description as input for recovering all semantics faithfully.}
    \label{fig:recon_vis}
    \vspace{-14pt}
\end{figure*}

\begin{figure*}[!t]
    \centering
    \includegraphics[width=0.90\linewidth]{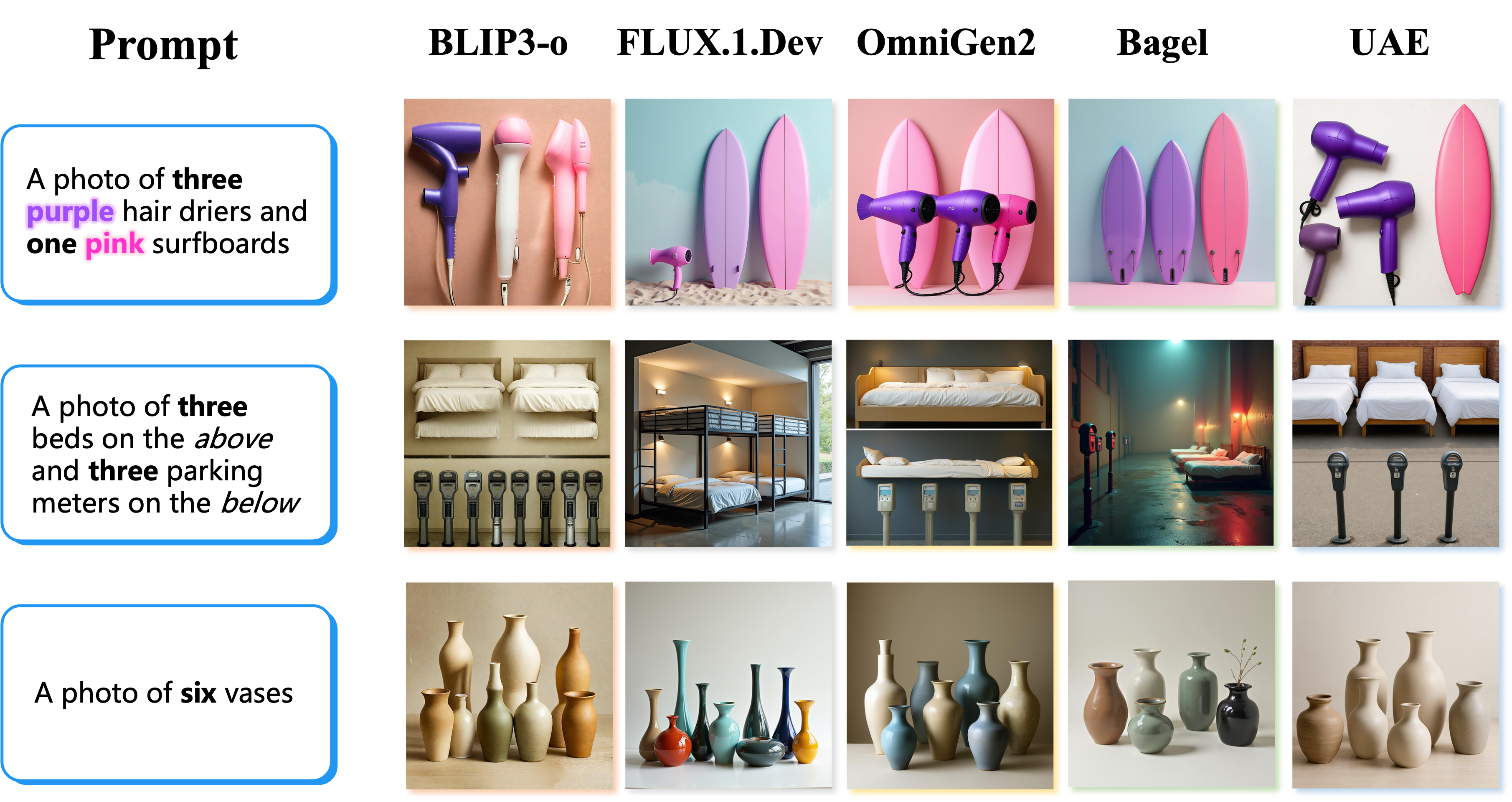}
    \vspace{-10pt}
    \caption{\textbf{Qualitative examples on GenEval++.} Under the complex and anti-realistic cases, our method demonstrates a clear advantage in multi-attribute instruction-following generation over other methods.}
    \label{fig:geneval_pp}
\end{figure*}

\begin{figure*}[!t]
    \centering
    \includegraphics[width=0.95\linewidth]{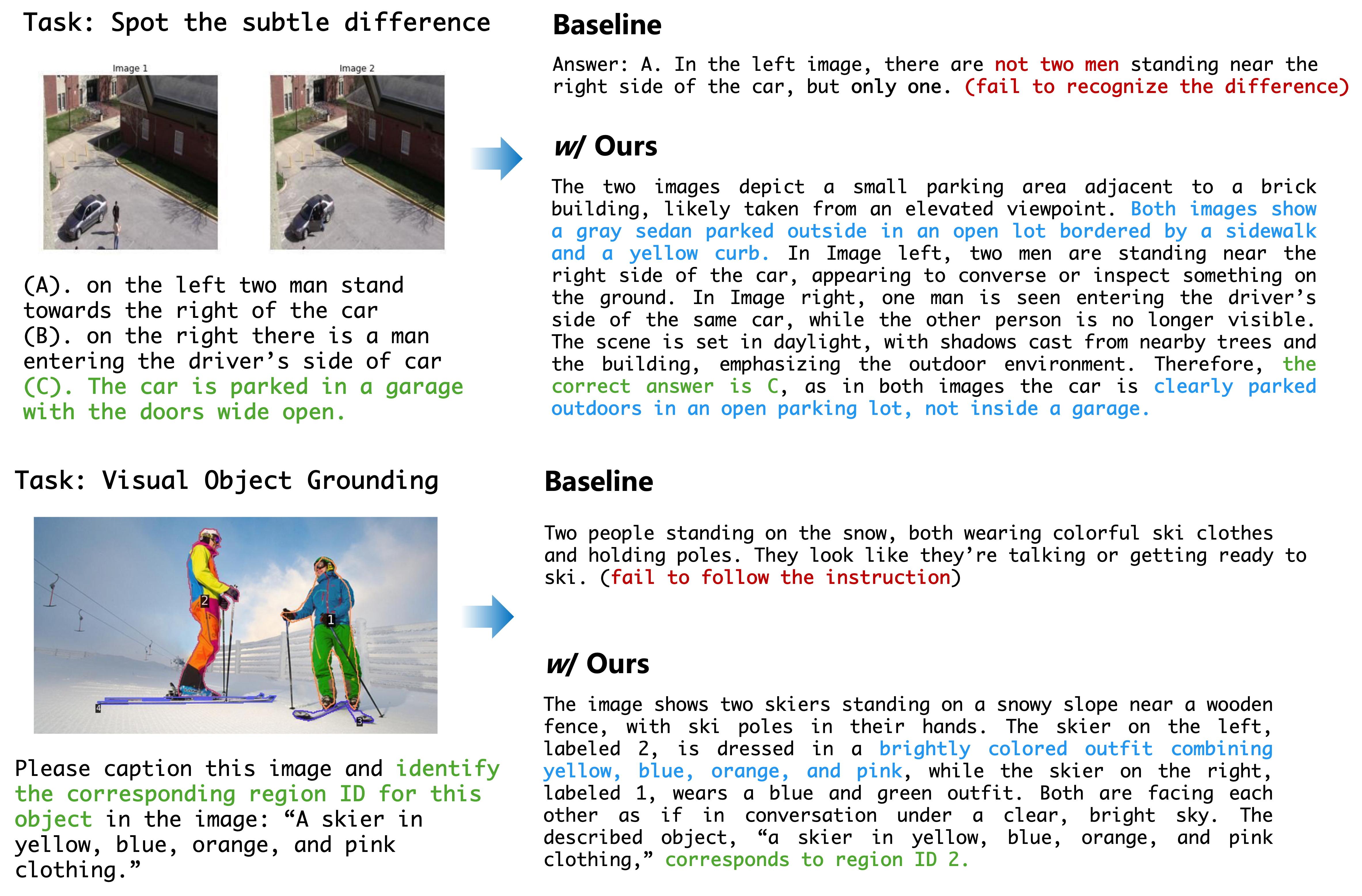}
    \vspace{-10pt}
    \caption{\textbf{Qualitative examples showing how reconstruction-driven RL improves image-to-text
understanding.} Compared to the baseline, our model better identifies subtle differences and performs accurate visual grounding, demonstrating that reconstruction-driven RL encourages richer and more precise semantic extraction in image-to-text understanding.}
    \label{fig:und_example}
\end{figure*}

\begin{table}[t]
    \centering{
    \renewcommand{\arraystretch}{1.3}
    \caption{Comparisons of \textbf{challenging instruction following generation ability} with other unified multimodal models on Geneval++~\citep{ghosh2024geneval}. \textbf{Bold} indicates the best result, and \underline{underlined} denotes the second best.}
    \vspace{-10pt}
    \resizebox{\linewidth}{!}{
    \setlength\tabcolsep{2pt}
        \begin{tabular}{lcccccccc}
            \toprule
            Method & \multicolumn{1}{c}{Color} & \multicolumn{1}{c}{Count} & \multicolumn{1}{c}{Color/Count} & \multicolumn{1}{c}{Color/Pos} & \multicolumn{1}{c}{Pos/Count} & \multicolumn{1}{c}{Pos/Size} & \multicolumn{1}{c}{Multi-Count} & \multicolumn{1}{c}{Overall}\\
            \midrule
            Janus-Pro~\citep{chen2025janus} & 0.450 & 0.300 & 0.125 & 0.300 & 0.075 & 0.350 & 0.125 & 0.246\\
            T2I-R1~\citep{jiang2025t2i} & \textbf{0.675} & 0.325 & 0.200 & 0.350 & 0.075 & 0.250 & 0.300 & 0.311 \\
            BLIP3-o 4B~\citep{chen2025blip3} & 0.125 & 0.225 & 0.100 & 0.450 & 0.125 & \underline{0.550} & 0.225 & 0.257\\
            BLIP3-o 8B~\citep{chen2025blip3} & 0.250 & 0.250 & 0.125 & \textbf{0.600} & 0.125 & \textbf{0.575} & 0.225 & 0.307 \\
            OmniGen2~\citep{wu2025omnigen2} & \underline{0.550} & 0.425 & 0.200 & 0.275 & 0.125 & 0.250 & \textbf{0.450} & 0.325 \\
            Bagel~\citep{bagel} & 0.325 & \underline{0.600} & \underline{0.250} & 0.325 & 0.250 & 0.475 & 0.375 & \underline{0.371} \\
            \midrule
            \rowcolor{blue!5}\textbf{\texttt{UAE}} & \underline{0.550} & 0.525 & \textbf{0.550} & \underline{0.550} & \textbf{0.450} & 0.400 & \underline{0.400} & \textbf{0.475} \\         
            \bottomrule
        \end{tabular}
    }
    \vspace{-5pt}
    \label{tab:Geneval++}
    }
\end{table}

\begin{table}[!t]
\scriptsize
    \centering
    \caption{\textbf{Protocol-2 of Unified-Bench}: evaluating the quality of output caption of our trained understanding model (3B) against different opponents on Unified-Bench, evaluated by four judge models (using official commercial API). We use the metric of \textbf{Pairwise winning rate (\%)} for evaluation. 
    The \textbf{Avg} column reports the mean score across judges. }
    \vspace{-10pt}
    \resizebox{\linewidth}{!}{
    \setlength{\tabcolsep}{3pt}
    \begin{tabular}{lcccccc}
        \toprule
        \multirow{2}{*}{Opponent} & \multirow{2}{*}{\# Param} & \multicolumn{5}{c}{Our Wining Rate (\%)}\\
        \cmidrule(lr){3-7} & & Claude-4.1 & GPT-4o & Grok-4 & o4-mini & Avg \\
        \midrule
        GPT-4o~\citep{gpt4o}      & - & 47.4 & 89.4 & 30.6 & 21.2 & 47.2 \\
        Bagel~\citep{bagel}       & 7B & 57.7 & 92.9 & 58.3 & 48.2 & 64.3 \\
        OmniGen2~\citep{wu2025omnigen2} & 3B & 67.9 & 97.6 & 63.5 & 56.5 & 71.4 \\
        Show-o~\citep{xie2024show}      & 1.3B & 97.8 & 100.0 & 89.8 & 91.0 & 94.7 \\
        \midrule
        Qwen-2.5-VL-3B~\citep{bai2025qwen25vl}  & 3B & 76.3 & 99.0 & 67.0 & 63.0 & 76.3 \\
        Qwen-2.5-VL-7B~\citep{bai2025qwen25vl}  & 7B & 68.8 & 99.0 & 62.0 & 56.0 & 71.5 \\
        \bottomrule
    \end{tabular}}
    \vspace{-2mm}
    \label{tab:unifiedbench_pairwise_winrate}
\end{table}


\section{Experiments}

\subsection{Ablation on Unified-GRPO}

To comprehensively evaluate the effectiveness of the proposed Unified-GRPO, we implement our method on the two typical unified multimodal models: UniWorld~\cite{lin2025uniworld} and Janus-Pro~\cite{chen2025janus}, among the understanding, generation, and unification benchmarks.
Tab.~\ref{tab:ablation} shows that applying Unified-GRPO to both representative UMM architectures consistently improves their performance. The gains are most significant on \textit{generation} and \textit{unification} metrics, where reconstruction is directly optimized, yielding improvements of $4\!\sim\!5\%$ on generation and over $6\%$ on unified reconstruction quality. 
Understanding performance exhibits only modest gains, which we attribute to the limited capacity of current generation models: imperfect reconstructions can introduce negative feedback to the encoder. Nevertheless, as we will show later (Sec.~\ref{mmt}), Unified-GRPO can notably enhance fine-grained perceptual abilities, particularly in tasks involving subtle difference recognition and visual grounding via our reconstructive RL training.
Since the UniWorld-based model demonstrates stronger performance in both generation and understanding compared to Janus, we adopt this architecture as the primary backbone for all subsequent experiments.

\subsection{Unification Evaluation}

\vspace{-2mm}

We assess the unified degree with the proposed Unified-Bench. Tab.~\ref{tab:unified-bench} shows that our \textbf{\texttt{UAE}} achieves the best \textbf{Overall} unified score (86.09), surpassing GPT\mbox{-}4o\mbox{-}Image (85.95). Specifically, \texttt{UAE} obtains the top results on CLIP (90.50), DINO-v2 (81.98), and DINO-v3 (77.54), and statistical parity on LongCLIP (94.35 vs.\ 94.37). These consistent gains across contrastive (CLIP-family) and self-supervised (DINO-family) features suggest that our \textbf{\texttt{UAE}} framework can preserve layout- and texture-level semantics that translate into more faithful reconstructions.

\begin{table*}[t!]
\centering
\caption{\textbf{High-level meta-tasks evaluation results on the comprehensive multimodal understanding benchmark:} MMT-Bench~\citep{mmt}. Accuracy is the metric, and the Overall score is computed as the mean of all displayed subtasks.}
\vspace{-10pt}
\resizebox{0.98\textwidth}{!}{%
\begin{tabular}{l|l|llllllllllllllll}
\toprule
Model & Overall & VR & Loc & Count & HLN & VC & VG & AR & PLP & I2IT & RR & Emo & VI & OCR & DU & IR & 3D \\
\midrule
Frequency Guess & 32.3 & 30.0 & 28.2 & 28.2 & 43.4 & 28.2 & 29.1 & 30.0 & 29.4 & 30.8 & 33.5 & 30.1 & 52.1 & 30.4 & 37.6 & 29.9 & 26.5 \\
Random Guess    & 27.9 & 27.1 & 28.1 & 25.0 & 41.6 & 25.0 & 24.8 & 26.6 & 21.2 & 33.4 & 10.5 & 25.4 & 50.8 & 27.2 & 30.3 & 24.3 & 25.5 \\
\midrule
InternVL-Chat-v1.2-34B & 58.7 & 81.3 & 59.4 & 66.4 & 82.4 & 82.3 & 49.4 & 52.6 & 37.4 & 32.8 & 55.0 & 48.7 & 61.5 & 60.5 & 68.3 & 56.3 & 45.5 \\
Qwen-VL-Plus            & 56.8 & 82.6 & 55.3 & 61.1 & 69.9 & 86.5 & 43.6 & 53.4 & 43.1 & 37.8 & 53.0 & 41.6 & 50.3 & 65.6 & 77.3 & 40.7 & 46.5 \\
GPT-4V                  & 54.1 & 85.3 & 55.6 & 51.6 & 69.6 & 80.3 & 25.0 & 47.7 & 48.2 & 31.8 & 52.5 & 45.1 & 47.9 & 68.0 & 69.8 & 44.9 & 42.0 \\
GeminiProVision         & 56.2 & 84.7 & 43.6 & 56.4 & 65.9 & 80.1 & 33.0 & 57.4 & 40.3 & 31.5 & 58.5 & 55.2 & 47.5 & 59.5 & 71.6 & 68.4 & 45.2 \\
DeepSeek-VL-7B          & 48.0 & 75.6 & 42.0 & 44.5 & 60.6 & 69.1 & 38.4 & 44.8 & 38.3 & 23.5 & 48.8 & 43.8 & 47.7 & 61.1 & 51.9 & 30.5 & 47.2 \\
Claude3V-Haiku          & 47.4 & 74.3 & 44.8 & 51.1 & 63.6 & 67.6 & 26.9 & 46.2 & 35.5 & 22.8 & 50.0 & 35.2 & 42.9 & 54.4 & 69.8 & 34.6 & 38.2 \\
ShareGPT4V-7B           & 47.8 & 74.2 & 36.0 & 50.9 & 62.4 & 71.6 & 35.4 & 46.2 & 39.2 & 21.8 & 59.8 & 44.3 & 54.5 & 47.8 & 47.9 & 27.8 & 45.2 \\
LLaVA-v1.5-7B           & 46.1 & 72.8 & 34.3 & 47.5 & 61.6 & 68.1 & 34.0 & 46.6 & 36.0 & 22.2 & 58.0 & 42.5 & 57.6 & 45.0 & 40.8 & 26.1 & 44.8 \\
\midrule
Qwen-2.5-VL-3B & 56.3 & 78.7 & 40.3 & 42.8 & 72.5 & 83.6 & 46.2 & 53.0 & 40.8 & 32.5 & 71.3 & 47.5 & 48.4 & 75.0 & 70.0 & 56.8 & 42.5 \\
Ours (Qwen-3B) & 56.5 & 80.1 & 47.3 & 44.7 & 72.8 & 84.1 & 47.1 & 53.5 & 46.6 & 32.7 & 71.3 & 48.3 & 57.6 & 68.8 & 58.4 & 50.6 & 40.0 \\
\rowcolor[HTML]{F9FFF9}
\textit{\textit{vs.} Baseline} & \bf {\textcolor[rgb]{0, 0.5, 0}{+0.2}} & \bf {\textcolor[rgb]{0, 0.5, 0}{+1.4}} & \bf {\textcolor[rgb]{0, 0.5, 0}{+7.0}} & \bf {\textcolor[rgb]{0, 0.5, 0}{+1.9}} & \bf {\textcolor[rgb]{0, 0.5, 0}{0.3}} & \bf {\textcolor[rgb]{0, 0.5, 0}{+0.5}} & \bf {\textcolor[rgb]{0, 0.5, 0}{+0.9}} & \bf {\textcolor[rgb]{0, 0.5, 0}{+0.5}} & \bf {\textcolor[rgb]{0, 0.5, 0}{+5.8}} & \bf {\textcolor[rgb]{0, 0.5, 0}{+0.2}} & \bf {\textcolor[rgb]{0.5, 0.5, 0.5}{+0.0}} & \bf {\textcolor[rgb]{0, 0.5, 0}{+0.8}} & \bf {\textcolor[rgb]{0, 0.5, 0}{+9.2}} & \bf {\textcolor[rgb]{0.5, 0, 0}{-6.2}} & \bf {\textcolor[rgb]{0.5, 0, 0}{-11.6}} & \bf {\textcolor[rgb]{0.5, 0, 0}{-6.2}} & \bf {\textcolor[rgb]{0.5, 0, 0}{-2.5}} \\
\bottomrule
\end{tabular}%
}
\vspace{-2mm}
\label{tab:overall-results}
\end{table*}

\begin{table*}[t]
\centering
\caption{\textbf{Evaluation results on fine-grained visual perception oriented sub-tasks} on MMT-Bench~\citep{mmt}. Accuracy is the metric, and the Overall score is computed as the mean of all displayed subtasks. We show notable improvements across various fine-grained understanding tasks, highlighting the positive impact of generation on understanding.}
\vspace{-10pt}
\resizebox{\textwidth}{!}{%
\begin{tabular}{l|c|ccccc|ccccc}
\toprule
 &  & \multicolumn{5}{c|}{Fine-grained Visual Recognition} & \multicolumn{5}{c}{Color and Geometry Perception} \\
\cmidrule(lr){3-7}\cmidrule(lr){8-12}
Model & Overall & \begin{tabular}[c]{@{}c@{}}Salient Obj.\\Detection RGBD\end{tabular} & \begin{tabular}[c]{@{}c@{}}Transparent\\Object Det.\end{tabular} & \begin{tabular}[c]{@{}c@{}}Small Object\\Detection\end{tabular} & \begin{tabular}[c]{@{}c@{}}Rotated Object\\Detection\end{tabular} & \begin{tabular}[c]{@{}c@{}}Person\\Re-ID\end{tabular} & \begin{tabular}[c]{@{}c@{}}Color\\Constancy\end{tabular} & \begin{tabular}[c]{@{}c@{}}Color\\Assimilation\end{tabular} & \begin{tabular}[c]{@{}c@{}}Geometrical\\Relativity\end{tabular} & \begin{tabular}[c]{@{}c@{}}Geometrical\\Perspective\end{tabular} & \begin{tabular}[c]{@{}c@{}}Polygon\\Localization\end{tabular} \\
\midrule
InternVL-Chat-V1.2-34B & 63.4 & 28.5 & 66.5 & 64.5 & 46.7 & 60.0 & 34.5 & 44.5 & 82.5 & 75.0 & 46.1 \\
Qwen-VL-Plus          & 62.3 & 44.5 & 47.5 & 59.5 & 60.0 & 50.5 & 47.5 & 29.0 & 58.3 & 43.0 & 63.8 \\
GPT-4V                & 62.0 & 42.0 & 56.5 & 52.0 & 79.0 & 49.0 & 65.0 & 24.7 & 43.3 & 35.7 & 66.0 \\
GeminiProVision       & 61.6 & 45.0 & 38.5 & 43.0 & 50.0 & 72.5 & 38.9 & 53.5 & 46.0 & 43.3 & 36.0 \\
DeepSeek-VL-7B        & 53.2 & 40.0 & 53.5 & 43.5 & 36.7 & 32.5 & 27.5 & 52.0 & 54.2 & 56.0 & 23.4 \\
Claude3V-Haiku        & 52.2 & 43.0 & 19.5 & 44.0 & 46.7 & 35.0 & 38.5 & 58.5 & 55.8 & 56.5 & 66.7 \\
ShareGPT4V-7B         & 51.5 & 40.5 & 39.0 & 37.5 & 27.8 & 24.0 & 52.8 & 26.5 & 60.0 & 65.8 & 32.0 \\
LLaVA-v1.5-7B         & 49.5 & 37.5 & 40.0 & 31.5 & 30.0 & 23.0 & 56.9 & 28.0 & 64.0 & 70.0 & 34.0 \\
Frequency             & 31.7 & 26.0 & 26.0 & 27.5 & 28.9 & 30.0 & 52.8 & 51.0 & 50.5 & 53.3 & 31.5 \\
Random                & 28.5 & 28.5 & 29.0 & 27.0 & 24.4 & 26.0 & 48.6 & 50.0 & 50.5 & 51.7 & 27.5 \\
\midrule
Qwen-2.5-VL-3B & 32.5 & 25.0 & 15.0 & 5.0 & 33.3 & 15.0 & 28.6 & 50.0 & 60.0 & 58.3 & 35.0 \\
Ours (Qwen-3B) & 56.9 & 45.0 & 45.0 & 45.0 & 55.6 & 75.0 & 42.9 & 60.0 & 65.0 & 75.0 & 60.0 \\
\rowcolor[HTML]{F9FFF9}
\textit{\textit{vs.} Baseline} & \bf {\textcolor[rgb]{0, 0.5, 0}{+24.4}} & \bf {\textcolor[rgb]{0, 0.5, 0}{+20}} & \bf {\textcolor[rgb]{0, 0.5, 0}{+30}} & \bf {\textcolor[rgb]{0, 0.5, 0}{+40}} & \bf {\textcolor[rgb]{0, 0.5, 0}{+22.3}} & \bf {\textcolor[rgb]{0, 0.5, 0}{+60}} & \bf {\textcolor[rgb]{0, 0.5, 0}{+14.3}} & \bf {\textcolor[rgb]{0, 0.5, 0}{+10}} & \bf {\textcolor[rgb]{0, 0.5, 0}{+5}} & \bf {\textcolor[rgb]{0, 0.5, 0}{+16.7}} & \bf {\textcolor[rgb]{0, 0.5, 0}{+25}} \\
\bottomrule
\end{tabular}%
}
\vspace{-5mm}
\label{tab:merged_selected_cols}
\end{table*}

\begin{figure*}[t]
    \centering
    \includegraphics[width=0.9\linewidth]{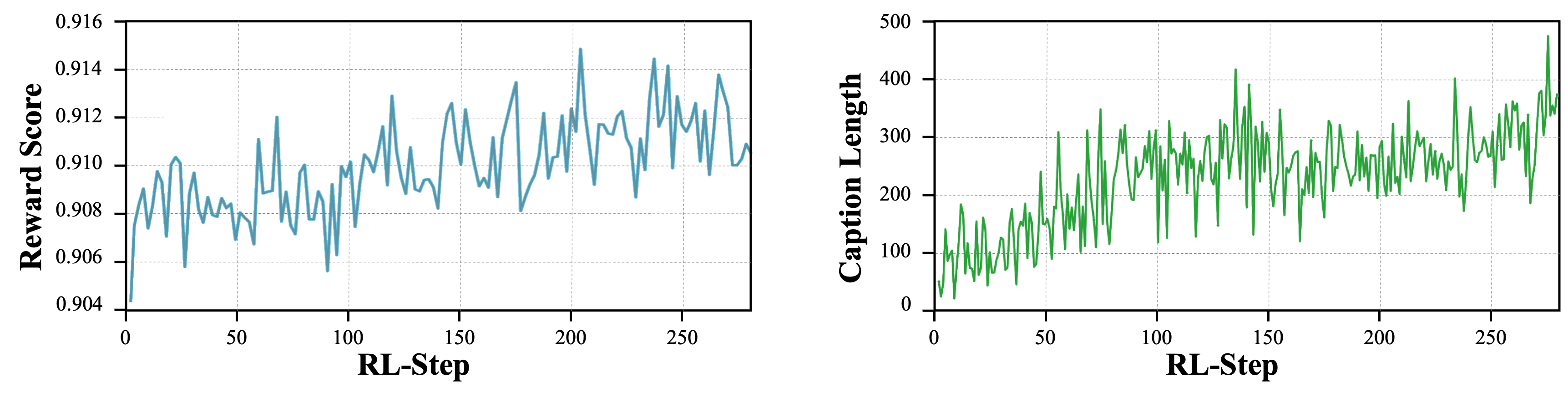}
    \vspace{-2mm}
    \caption{
    \textbf{Training dynamics of our reconstruction-oriented RL stage. }
    \textbf{Left:} The reward score steadily increases as the policy learns to generate captions that more faithfully preserve the visual information in the input image. 
    \textbf{Right:} The caption length gradually grows throughout training, indicating that the model is producing richer and more detailed textual descriptions. 
    Together, these trends show that the RL optimization encourages the model to encode progressively more complete image information into text, ensuring that the downstream decoder receives a maximally informative representation.
    }
    \label{fig:longer}
    \vspace{-5mm}
\end{figure*}

\subsection{Text-to-Image Generation Evaluation}

We evaluate \textbf{\texttt{UAE}} on two standard benchmarks: GenEval and its improved version GenEval++, which probe compositional understanding and instruction-following in increasingly challenging settings.
More evaluations are in the Supplementary.

\textbf{GenEval.} As shown in Tab.~\ref{tab:geneval}, without considering LLM rewriting, our \textbf{\texttt{UAE}} attains the best \emph{Overall} score among unified models (\textbf{0.86}). It leads on \emph{Counting} (0.84) and \emph{Color attribution} (0.79; +\textbf{16} points vs.\ Bagel’s 0.63 and +3 vs.\ OmniGen2’s 0.76), co-leads on \emph{Colors} (0.90), is second-best on \emph{Position} (0.71), and reaches 0.89 on \emph{Two object} (below the strongest 0.94–0.95). When considering LLM rewriting, e.g., using the same rewritten prompts with Bagel, our \texttt{UAE} achieves an overall score of 0.89 on average, demonstrating the SOTA performance in image generation.

\textbf{GenEval++ (harder compositional control).} GenEval++~\citep{ye2025echo} extends GenEval to prompts with \emph{three or more} objects, each bearing distinct attributes and spatial relations, demanding comprehensive, multi-constraint satisfaction. In Tab.~\ref{tab:Geneval++}, \textbf{\texttt{UAE}} achieves the best \emph{Overall} score (\textbf{0.475}), leading on \emph{Color/Count} (0.550) and \emph{Pos/Count} (0.450), with runner-up performance on \emph{Color/Pos} (0.550) and \emph{Multi-Count} (0.400). Qualitative visualizations in Fig.~\ref{tab:Geneval++} further show accurate attribute binding, disambiguation across multiple entities, and robust position–count consistency under long prompts.
This highlights that our method can achieve notable improvement in complex instruction following.


\subsection{Image-to-Text Understanding Evaluation}
\label{sec:improved-understanding}

Here, we conduct experiments to verify that after our post-training, the encoder can achieve improved image-to-text, in terms of caption quality, and greater ``generation-friendly".

\noindent\textbf{Caption quality evaluation by commercial LLMs.}
As shown in Tab.~\ref{tab:unifiedbench_pairwise_winrate}, our understanding model (using Qwen-2.5-VL-3B as the baseline) attains high average win rates: \textbf{94.7} vs.\ Show-o, \textbf{71.4} vs.\ OmniGen2, \textbf{64.3} vs.\ Bagel, and \textbf{76.3}/\textbf{71.5} vs.\ Qwen-2.5-VL (3B/7B), while remaining competitive with GPT-4o (47.2). The cross-judge agreement suggests our captions improve along multiple axes, completeness, attribute binding, relational, and spatial fidelity.

\noindent\textbf{Improving the understanding model as a better captioner suitable for generation.}
Under the Unified-Bench "caption$\to$generate$\to$compare" protocol, captions produced by our trained understanding model yield the highest reconstruction similarity across all four backbones (Tab.~\ref{tab:captioner_for_gen}): \textbf{90.50} (CLIP), \textbf{94.35} (LongCLIP), \textbf{81.98} (DINO-v2), \textbf{77.54} (DINO-v3), with \textbf{86.09} Overall.
These results indicate that the caption generated by our understanding model is more suitable for generation.

\subsection{Evaluation on the Understanding Benchmark.} 
\label{mmt}
We evaluate on \emph{MMT-Bench}~\citep{mmt}, which comprises high-level meta-tasks\footnote{The tasks include VR (Visual Recognition), Loc (Spatial Localization), OCR (Text Reading), Count (Object Counting), HLN (Hallucination), IR (Image Retrieval), 3D, VC (Visual Caption), VG (Visual Grounding), DU (Document Understanding), AR (Action Recognition), PLP (Pixel-Level Perception), I2IT (Image-to-Image Translation), RR (Relation Reasoning), Emo (Emotion), and VI (Visual Illusion).}
The overall score remains essentially unchanged with a marginal improvement over the baseline (+0.2\%; Tab.~\ref{tab:overall-results}). However, if we zoom in to observe \emph{fine-grained visual recognition} suite (Tab.~\ref{tab:merged_selected_cols}), the benefits of our generation-augmented training for perception become pronounced: we observe large absolute gains in Small Object Detection (+40.0\%) and Person Re-ID (+60.0\%), yielding a +24.4\% increase in the fine-grained overall. 
These results indicate that generation does not harm understanding, but can instead \textbf{enhance fine-grained visual perception capability.}

\subsection{Case-Study in Fine-Grained Visual Perception}

\paragraph{Spotting Subtle Differences.}
Fig.~\ref{fig:und_example} (top) presents a challenging visual comparison task, where the baseline model fails to detect the fine difference between two images: one image
shows \emph{two} men standing near the car, while in the other image \emph{only one} man is visible. The baseline incorrectly answers option A due to missing the subtle change.
In contrast, our model, trained with Unified-GRPO, provides a detailed analysis of both
images, accurately recognizing the presence and position of each person, the vehicle’s
location, the outdoor parking setting, and the contextual cues.

\vspace{-2mm}

\paragraph{Visual Object Grounding.}
Fig.~\ref{fig:und_example} (bottom) demonstrates another demanding task requiring precise grounding. The input instruction asks the model to caption the scene \emph{and identify the corresponding region ID} for “a skier in yellow, blue, orange, and pink clothing.” The baseline generates a generic caption and completely ignores the grounding instruction.
After our training, the model not only follows the instruction faithfully but also grounds the described skier to the correct region ID by identifying the color composition of the outfit and matching it to the labeled bounding box.

\vspace{-1mm}
\section{Related Work}
\vspace{-1mm}

Recent advancements in multimodal AI have led to the development of Unified Multimodal Models (UMMs)~\cite{zhang2025unified}. The architectural designs of current UMMs can be broadly categorized into two paradigms:
\textbf{(1) AR-based Approaches}: In this setup, all modalities, including images and text, are tokenized and processed sequentially using an autoregressive transformer. Systems like Chameleon and EMU generate image tokens akin to language modeling by predicting the next token in a sequence~\cite{team2024chameleon,qu2024tokenflow,wu2025janus,chen2025janus,liquid,freestyleret}. 
An evolution of this idea is seen in Show-o~\cite{xie2024show}, which enhances token prediction with a discrete diffusion mechanism, introducing a structured denoising process during generation.
\textbf{(2) Hybrid AR-Diffusion Architectures:} Some models combine autoregressive modeling with diffusion-based image synthesis~\cite{yan2025gpt}. For instance, Transfusion and similar systems~\cite{zhou2024transfusion,bagel,ma2024janusflow,shi2024llamafusion,xie2025show} extend a shared transformer backbone with a dedicated diffusion or flow-matching head for high-fidelity image generation. Alternatively, other approaches freeze a pre-trained MLLM and use learnable query modules or MLPs to extract and route intermediate representations to an external image generator~\cite{pan2025metaquery,chen2025blip3,lin2025uniworld}.
A more recent direction integrates standard autoregressive language processing with masked-autoregressive reconstruction for visual data. MAR~\cite{li2024autoregressive} enables image generation without relying on vector quantization, instead reconstructing patches in a flexible order. This approach has been adopted in models such as Harmon~\cite{wu2025harmon,fan2025unified,wang2025skywork}. Meanwhile, some works~\cite{geng2025x,chen2025blip3} use a discretized SigLIP~\cite{tschannen2025siglip} to convert images into tokens, training a single autoregressive model over these visual and language tokens, while employing a diffusion model for the final image decoding.
Similar post-training works~\cite{wang2025visual,xie2025reconstruction} based on reconstruction demonstrate that using the dense image feature as the ``rich text" condition for training diffusion models, which improves image generation. Additionally, RL-based frameworks have been proposed to enhance multimodal learning~\cite{liu2025flow,xue2025dancegrpo,xu2023imagereward,piao2026towards}.


\section{Conclusion}
\label{sec:conclusion}



We show that an auto-encoder can serve as a foundational architecture for unifying image-to-text understanding and text-to-image generation.
This paradigm leverages text as a shared intermediate latent representation.
By introducing Unified-GRPO, we jointly optimize both, creating a synergistic feedback loop, enabling the auto-encoder principle to benefit both understanding and generation tasks simultaneously.
This simple yet powerful design yields stronger fine-grained visual perception, richer semantic encoding, and improved complex instruction-following capability. 
Our findings highlight the value of treating multimodal tasks not as isolated objectives but as mutually reinforcing components of a unified system, paving the way for more coherent and synergistic multimodal learning.

\clearpage

\paragraph{Acknowledgment.}
This work was supported in part by the Natural Science Foundation of China (No. 62332002, 62425101), The Guangdong Grants (Grant No.2023ZT10X075), and Shenzhen Science and Technology Program (KQTD20240729102051063).




{
    \small
    \bibliographystyle{ieeenat_fullname}
    \bibliography{main}
}

\clearpage
\setcounter{page}{1}
\maketitlesupplementary
\setcounter{section}{0}

\section*{Supplementary Overview}
\begin{itemize}
    \item Section~\ref{appendix:related}: Additional Related works.
    \item Section~\ref{appendix:dataset}: Dataset details.
    \item Section~\ref{appendix:training}: Training settings.
    \item Section~\ref{appendix:qualitative}: Qualitative examples.
    \item Section~\ref{appendix:exp_details}: Additional experimental results.
\end{itemize}

\section{Additional Related works}
\label{appendix:related}

\paragraph{Reinforcement Learning in Generative Models.}

The widespread success of Reinforcement Learning from Human Feedback (RLHF) in aligning large language models (LLMs) with human intent~\cite{christiano2017deep,hu2022lora} has inspired its application to text-to-image generation. 
In this context, a common strategy involves first training a reward model (RM) that learns from human judgments—either general aesthetic preferences~\cite{xu2024visionreward} or alignment between prompts and generated images~\cite{xu2023imagereward}, followed by reinforcement learning to optimize the generative model accordingly~\cite{black2023training}. Despite its promise, this two-stage approach faces significant limitations when applied to image editing tasks. Reward models are often brittle and challenging to design robustly~\cite{miao2024training}, and they can be gamed through superficial changes that maximize reward without improving actual quality, a phenomenon known as "reward hacking"~\cite{wang2025pref}.
More recently, alternative optimization frameworks like GRPO~\cite{shao2024deepseekmath} have emerged as viable solutions, demonstrating effectiveness in tuning both diffusion and flow-matching based models. Extensions such as FlowGRPO~\cite{liu2025flow} and DanceGRPO~\cite{xue2025dancegrpo} illustrate the adaptability of these algorithms to complex generative processes, offering a more stable and fine-grained path toward aligning visual outputs with human expectations—particularly in dynamic, iterative editing scenarios where traditional methods fall short.

\paragraph{Benchmarking Multimodal Understanding, Generation, and Unification.}
Evaluating unified multimodal models (UMMs) typically involves aggregating performance across multiple specialized benchmarks, each targeting distinct capabilities. For assessing visual understanding, widely adopted benchmarks include ScienceQA~\cite{lu2022learn}, MMMU~\cite{yue2024mmmu}, VQA~\cite{antol2015vqa}, GQA~\cite{hudson2019gqa}, and MM-Bench~\cite{liu2024mmbench}, all of which rely heavily on large-scale datasets with human-annotated images and labels. In contrast, our proposed UniBench introduces a novel paradigm as a VQA-style benchmark specifically designed for \textit{generated} images, eliminating the dependency on real-image annotations by evaluating comprehension directly on synthesized content.
For generative capability assessment, image quality is commonly measured using metrics such as FID~\cite{heusel2017gans}, ImageReward~\cite{xu2023imagereward}, and LIQ~\cite{tian2025quality}, often evaluated on standard image corpora like MSCOCO~\cite{lin2014microsoft} or LAION-5B~\cite{schuhmann2022laion}. Additional factors such as text-image alignment~\cite{hessel2021clipscore}, fairness~\cite{lee2023holistic}, and stylistic consistency~\cite{peng2024dreambench++} are also considered, drawing from benchmarks like HRS~\cite{bakr2023hrs}.
However, unified models place greater emphasis on instruction-following and coherent joint reasoning across perception and generation. As such, evaluation frameworks tailored to text-to-image synthesis, such as GenEval~\cite{ghosh2023geneval}, DPG-Bench~\cite{hu2024ella}, and T2I-CompBench++~\cite{huang2025t2i}, which are particularly relevant. These assess fine-grained attributes including object presence, spatial relations, counting accuracy, color fidelity, and positional reasoning~\cite{bakr2023hrs,li2024evaluating,cho2024davidsonian}. Despite their utility, existing benchmarks are not specifically designed for the dual perception-generation nature of UMMs, leaving a gap in comprehensive, integrated evaluation.
To address world-knowledge grounding in image synthesis, WISE~\cite{wisebench} was recently introduced to evaluate models' implicit understanding of real-world constraints across domains such as food preparation, material physics, and object affordances. More recently, UniEval~\cite{li2025unieval} proposes a new benchmark dedicated to unified multimodal modeling, covering a broader range of semantic, structural, and logical challenges with increased task difficulty and potential for model improvement.

\section{Dataset Details}
\label{appendix:dataset}

\noindent\textbf{RL stage data (1K).}
For the reinforcement learning (RL) phase, we curate a compact but highly refined dataset of 1,000 real-world photography images selected for exceptional compositional quality, visual clarity, and semantic richness. These images span diverse domains such as portrait photography, architectural shots, nature scenes, and dynamic street photography, all captured under realistic lighting and perspective conditions. In addition to these hand-picked photographs, we incorporate a specialized subset of synthetic yet photorealistic data from Echo-4o~\cite{ye2025echo}, which provides tightly aligned text-image pairs with expert-level captions and controlled visual variations. This combined RL dataset is used in a reconstruction-driven optimization framework: given a caption derived from one of these target images, the model is tasked with generating a new image, and its output is evaluated against the original using a learned reward model that assesses fidelity, detail preservation, and semantic alignment. Through this closed-loop paradigm, improved captioning leads to better reconstruction, which in turn refines generation capabilities.

\noindent\textbf{Data for evaluation in Unified-Bench.}
To evaluate the model's performance on the proposed Unified-Bench, we randomly sample 100 images from the LAION-5B dataset~\cite{schuhmann2022laion} to serve as a dedicated test split. These images are selected without any filtering or curation based on content or aesthetic score, ensuring a representative and unbiased distribution across categories, styles, and complexity levels.

\section{Training Settings}
\label{appendix:training}

\begin{figure}[t]
    \centering
    \includegraphics[width=0.95\linewidth]{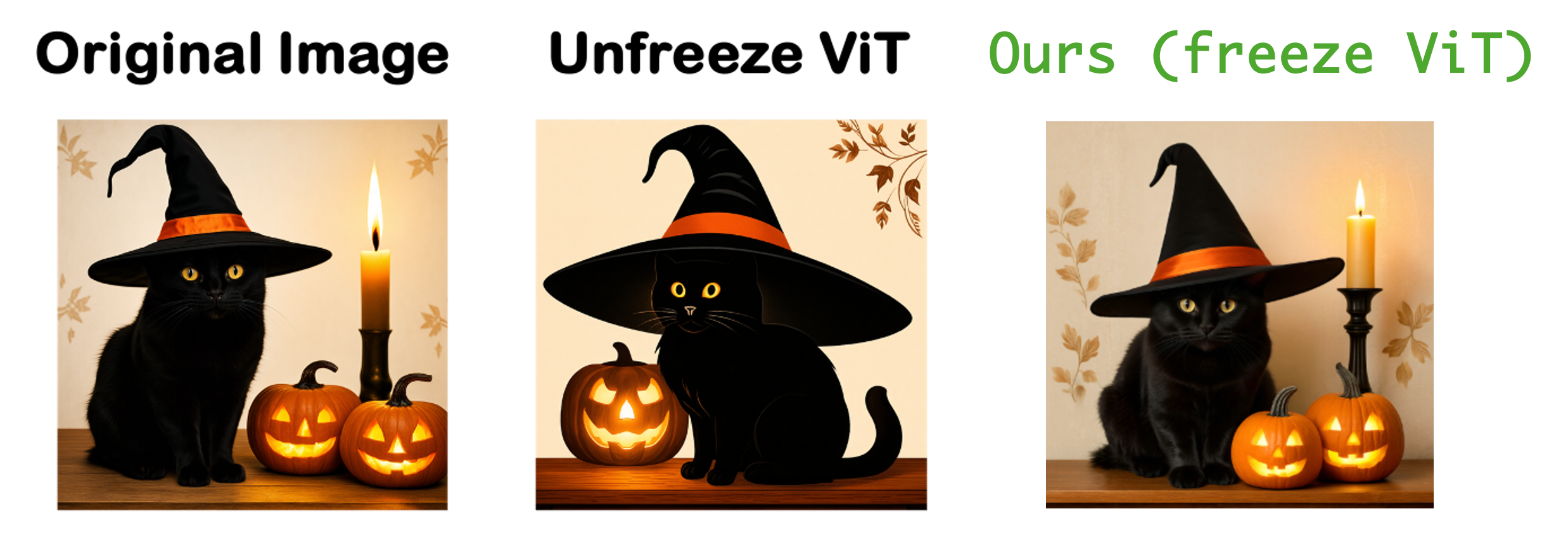}
    \caption{Illustration of the reconstruction results when unfreezing ViT (the visual encoder of the MLLM) for joint training. We observe that the generated output collapses, semantically important details such as ``two pumpkins" and ``one candle" are missing. This degradation motivates us to keep the ViT frozen during finetuning across all experiments.
    }
    \label{fig:vit}
\end{figure}

\paragraph{Training details of Unified-GRPO.}
We employ the GRPO RL algorithm~\cite{shao2024deepseekmath} to fine-tune only the LLM module while keeping other modules, like the corresponding visual encoder/decoder, frozen. 
We empirically observe that updating the visual encoder during RL training can lead to instability and degradation in image quality (see Fig.~\ref{fig:vit}), such as anomaly artifacts, structural collapse, or semantic inconsistency, so we disable its gradient updates to preserve visual feature integrity. 
To enable effective sampling for RL-based image generation, we treat the combination of the diffusion decoder and a pre-trained CLIP model~\cite{radford2021learning} as a unified, frozen reward module. This composite model operates purely in inference mode: given a generated image and its corresponding reconstructed image from the input caption, it computes a similarity score that serves as the final reward signal in the GRPO framework. 
During training, we use a learning rate of $1 \times 10^{-6}$ and a batch size of 1 due to the high computational cost of diffusion-based RL. For each prompt, we generate 4 sampled images to estimate the policy gradient in GRPO, and we set the KL regularization coefficient $1 \times 10^{-6}$, indicating that we only apply a minimal penalty for divergence from the reference policy, focusing solely on reward maximization. The temperature of LLM is set to be 1.0. The prompt used to do the LLM sampling is shown below (see Prompt.~\ref{llm_prompt}).

Note that we do \textbf{not} explicitly require the LLM to generate descriptive or comprehensive captions during training. After RL, the LLM autonomously produces longer and richer captions that are more conducive to high-fidelity image generation, even though no explicit supervision or loss is applied to the caption content itself. This emergent behavior suggests that the \textbf{\emph{RL signal from image reconstruction quality implicitly guides the LLM toward generating more detailed and image-friendly textual descriptions.}}


\section{Qualitative Examples}
\label{appendix:qualitative}

\begin{figure*}[t]
    \centering
    \includegraphics[width=\linewidth]{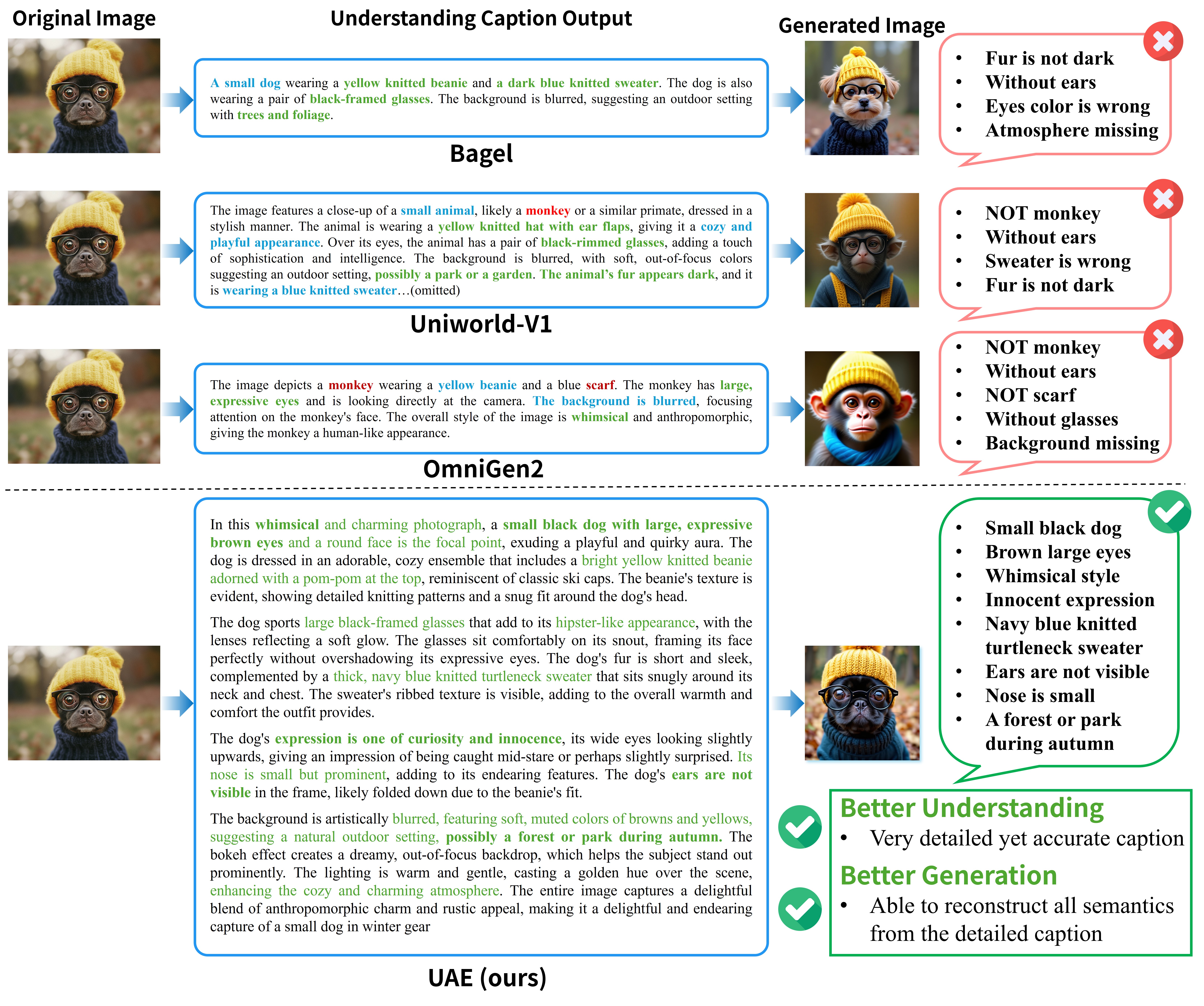}
    \caption{\textbf{Case study of the results from the proposed Unified-Bench}, we see that \textit{our UAE enables to produce a more detailed, accurate, comprehensive description based on the input image}, and reconstructs a similar result to the original image, showcasing the improved understanding and generation capabilities, and the better unification of the system.}
    \label{fig:case_study}
    \vspace{-2mm}
\end{figure*}

\paragraph{Enhancing model's comprehensive perception by the generation model.}
Fig.~\ref{fig:case_study} contrasts captions used for reconstruction on a challenging example (small black dog wearing a yellow beanie and glasses). Baselines reveal three typical errors. (i) \emph{Category drift}: some misidentify the subject as a monkey, causing the generator to synthesize an incorrect species. (ii) \emph{Attribute omissions or swaps}: descriptions drop key items (beanie, glasses) or mismatch apparel colors, leading to reconstructions that caricature the outfit. (iii) \emph{Under-specified scenes}: vague backgrounds and missing lighting cues prevent consistent photographic style at inference. UAE’s caption, in contrast, enumerates the full set of semantics—species, apparel \emph{type and color}, eyewear, pose, occlusions (“ears are not visible”), background style (“blurred, park-like”), and lighting—producing a reconstruction that preserves identity, attire, and overall aesthetic. This example typifies the mechanism by which better understanding (denser, better-bound descriptions) yields better generation, echoing our Unified-Bench gains in Tab.~\ref{tab:captioner_for_gen}.

\begin{figure*}
    \centering
    \includegraphics[width=0.80\linewidth]{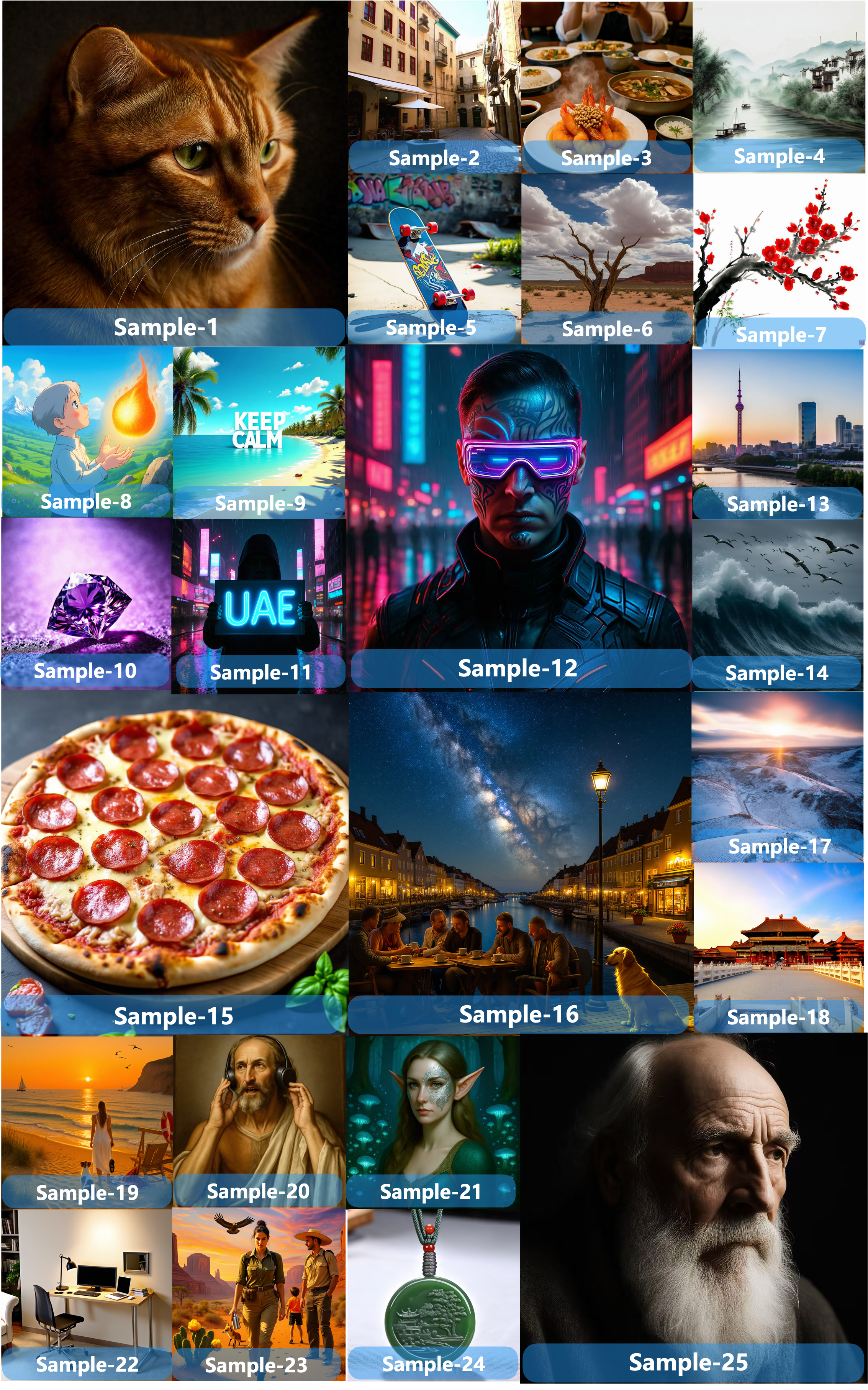}
    \caption{Visualization results of \texttt{UAE} at 1024$\times$1024 resolution.}
    \label{fig:vis}
\end{figure*}


\section{Additional Experimental Results}
\label{appendix:exp_details}

\paragraph{The text-to-image generation results on DPG-Bench.}

On DPG-Bench (Tab.~\ref{tab:dpgbench}), UAE achieves the top scores on \emph{Entity} (91.43), \emph{Attribute} (91.49), and \emph{Relation} (92.07), and ranks second overall with \textbf{84.74}, closely trailing Bagel (85.07). The sub-score pattern suggests UAE’s advantages come from faithful entity grounding and relation handling under long prompts, translating into competitive end-to-end generation quality within a unified architecture.

\begin{table*}[t]
    \centering
    \caption{Comparisons of text-to-image generation ability on DPG-Bench~\cite{hu2024ella} benchmark. \textbf{Bold} indicates the best result, and \underline{underlined} denotes the second best.}
\resizebox{1.0\textwidth}{!}{
\setlength\tabcolsep{10pt}
    \begin{tabular}{lccccccc}
        \toprule
        Method & \multicolumn{1}{c}{Global} & \multicolumn{1}{c}{Entity} & \multicolumn{1}{c}{Attribute} & \multicolumn{1}{c}{Relation} & \multicolumn{1}{c}{Other} & \multicolumn{1}{c}{Overall} \\
        \midrule
        \multicolumn{7}{c}{\textbf{Dedicated T2I}} \\
        \midrule
        SDXL~\cite{podell2023sdxl} & 83.27 & 82.43 & 80.91 & 86.76 & 80.41 & 74.65 \\ 
        PlayGroundv2.5~\cite{li2024playground} & 83.06 & 82.59 & 81.20 & 84.08 & 83.50 & 75.47 \\
        Hunyuan-DiT~\cite{li2024hunyuan} & 84.59 & 80.59 & 88.01 & 74.36 & 86.41 & 78.87 \\
        PixArt-$\Sigma$~\cite{chen2023pixart} & 86.89 & 82.89 & 88.94 & 86.59 & 87.68 & 80.54 \\
        DALLE3~\cite{dalle3} & \textbf{90.97} & 89.61 & 88.39 & 90.58 & 89.83 & 83.50 \\
        SD3-medium~\cite{sd3-medium} & 87.90 & \underline{91.01} & 88.83 & 80.70 & 88.68 & 84.08 \\
        FLUX.1-dev~\cite{FLUX} & 82.1 & 89.5 & 88.7 & \underline{91.1} & 89.4 & 84.0 \\ 
        OmniGen~\cite{xiao2025omnigen} & 87.90 & 88.97 & 88.47 & 87.95 & 83.56 & 81.16 \\
        \midrule
        \multicolumn{7}{c}{\textbf{Unified Model}} \\
        \midrule
        Show-o~\cite{xie2024show} & 79.33 & 75.44 & 78.02 & 84.45 & 60.80 & 67.27 \\
        EMU3~\cite{wang2024emu3} & 85.21 & 86.68 & 86.84 & 90.22 & 83.15 & 80.60 \\
        TokenFlow-XL~\cite{qu2025tokenflow} & 78.72 & 79.22 & 81.29 & 85.22 & 71.20 & 73.38 \\ 
        Janus Pro~\cite{chen2025janus} & 86.90 & 88.90 & 89.40 & 89.32 & \underline{89.48} & 84.19 \\
        BLIP3-o 4B~\cite{chen2025blip3} & - & - & - & - & - & 79.36 \\
        BLIP3-o 8B~\cite{chen2025blip3} & - & - & - & - & - & 81.60 \\
        UniWorld-V1~\cite{lin2025uniworld} & 83.64 & 88.39 & 88.44 & 89.27 & 87.22 & 81.38 \\
        OmniGen2~\cite{wu2025omnigen2} & 88.81 & 88.83 & 90.18 & 89.37 & \textbf{90.27} & 83.57 \\
        BAGEL~\cite{bagel} & \underline{88.94} & 90.37 & \underline{91.29} & 90.82 & 88.67 & \textbf{85.07} \\
        \rowcolor{blue!5}\textbf{UAE} & 83.11 & \textbf{91.43} & \textbf{91.49} & \textbf{92.07} & 84.32 & \underline{84.74} \\
        \bottomrule
    \end{tabular}
    }
\label{tab:dpgbench}
\end{table*}

\paragraph{Prompt list used in Fig.~\ref{fig:vis}.}

We provide the full caption for each sample in generation order, reading from left to right and top to bottom, row by row.

\begin{itemize}[nolistsep, leftmargin=*]
    \item \textbf{Sample-1}. A close-up portrait of a ginger tabby cat, its fur a rich tapestry of warm amber and deep russet stripes that catch the soft, directional light illuminating its face from the side, highlighting the velvety texture of its coat and the subtle contours of its cheekbones, while its large, luminous green eyes gaze intently off-camera with an expression of quiet contemplation and alert curiosity, framed by long, delicate white whiskers and perked ears that suggest attentiveness, all set against a dark, shadowy background that isolates the feline subject and enhances the dramatic, almost painterly quality of the image, emphasizing the cat’s regal poise and enigmatic presence.
    
    \item \textbf{Sample-2}. The building on the left is a light beige color with a series of rectangular windows framed in red, some with small white panes. These windows have simple brick or mortar surrounds and are uniformly spaced, creating a rhythmic pattern across the facade. The ground floor features a small shop area with a white canopy providing shade for outdoor seating. The canopy is supported by metal poles and holds a few tables under its shelter. Behind this canopy, various items can be seen, including a few chairs and tables, indicating a café or small eatery. A white umbrella stands next to the shop entrance, adding to the cozy atmosphere.Above the shop, the building has a series of small balconies with metal railings, each adorned with potted plants and hanging baskets, contributing to the pedestrian-friendly urban design. The ground floor has a mix of business signs, some of which are partially visible but not legible, suggesting a bustling commercial area. There's a dark green signboard affixed to one of the windows, possibly indicating a specialty shop or restaurant. The neighboring building on the right is a lighter shade of beige with a pastel green section near the top. Its windows are similarly framed in red, with larger panes and a more varied arrangement compared to the first building. This building features balconies with metal railings and small rectangular windows. The exterior walls show some wear and tear, with subtle moldings and patches of weathering, adding character to the structures. In front of these buildings lies a cobblestone street, which is partially shaded by the shadows cast by the buildings. A large stone fountain occupies the foreground, its base circular and gray, with a worn, dark surface. The pavement around the fountain is paved with irregularly shaped stones, creating a rustic, old-world feel. The sunlight creates dramatic contrasts, with deep shadows and bright highlights accentuating the textures of the buildings and the cobblestones. The street is quiet, devoid of people, which enhances the serene and timeless atmosphere of the scene. 

    \item \textbf{Sample-3}. A photo of hearty Chinese meal.

    \item \textbf{Sample-4.} This serene watercolor painting evokes the tranquil spirit of a traditional Chinese riverside village, where mist-laden mountains recede into a soft, pale sky, their layered silhouettes rendered in gentle washes of gray and muted green that dissolve into atmospheric haze; along the calm, reflective riverbank, white-walled houses with dark-tiled, upturned eaves nestle among lush trees, their architecture echoing classical Jiangnan aesthetics, while two slender wooden boats glide silently on the glassy water—one closer to the foreground with its simple mast and open cabin, the other a distant speck fading into the fog—imbuing the scene with quiet movement and timeless stillness, as the interplay of light and shadow across the rippling surface and the subtle gradations of ink suggest not only depth and distance but also a meditative harmony between nature and human habitation, capturing the essence of poetic rural life suspended in a dreamlike, almost ethereal moment.

    \item \textbf{Sample-5}. A vibrant blue skateboard with bold, graffiti-style graphics—featuring swirling red and yellow patterns and stylized lettering—stands upright on cracked concrete, its bright red wheels and silver trucks catching the sunlight, casting a sharp shadow on the ground, while in the blurred background, a weathered wall adorned with colorful street art and a partially visible skate ramp hint at an urban skate park setting, blending raw energy with artistic expression under a clear, sunlit sky.

    \item \textbf{Sample-6}. A solitary, gnarled tree with twisted, leafless branches stretches skyward like a skeletal sentinel in the heart of a vast desert landscape, its weathered trunk rooted firmly in the ochre sands that stretch to the horizon, dotted sparsely with low-lying shrubs; above, a dramatic expanse of billowing cumulus clouds drifts across a brilliant blue sky, casting shifting shadows over the arid terrain, while in the distance, the imposing silhouette of red rock mesas rises majestically against the horizon, lending a sense of ancient grandeur and timeless solitude to the scene, where nature’s raw resilience and stark beauty are captured in perfect harmony under the vast, open heavens.

    \item \textbf{Sample-7}. A striking traditional East Asian ink painting captures the vibrant essence of a blossoming plum tree, its gnarled, darkly rendered branches—executed with bold, expressive brushstrokes of sumi ink—arching gracefully across the stark white paper to cradle clusters of vivid crimson flowers, each petal delicately shaped with fluid washes of red that convey both vitality and fragility, while subtle hints of green foliage at the lower left suggest the quiet emergence of new life; the composition balances dynamic movement with serene stillness, evoking themes of resilience and renewal as the blossoms defiantly bloom against the void, enhanced by the faint calligraphic inscription near the trunk and the small red seal in the corner, which together anchor the piece in cultural tradition and artistic intention.

    \item \textbf{Sample-8}. In a breathtaking, sun-drenched meadow of lush rolling hills dotted with wildflowers and scattered boulders, a young boy with soft silver-gray hair and wide, awestruck blue eyes gazes upward in wonder as he gently cradles a radiant, living flame between his outstretched palms—a glowing, teardrop-shaped orb of golden-orange fire that pulses with warmth and light, its edges flickering with delicate embers against the backdrop of a brilliant blue sky streaked with fluffy white clouds and distant snow-capped mountains; dressed in a simple light-blue jacket over a crisp white shirt, the child embodies innocence and quiet awe, as if he has just summoned or discovered this mystical force, transforming the idyllic pastoral landscape into a realm where magic feels not only possible but tenderly held, evoking a sense of harmony between nature, wonder, and the boundless imagination of youth.

    \item \textbf{Sample-9}. A vibrant, sun-drenched tropical beach unfolds under a brilliant azure sky dotted with fluffy white clouds, where the crystal-clear turquoise waters gently lap against golden sands lined with swaying palm trees casting dappled shadows on the shore, and at the heart of this serene paradise, the bold, three-dimensional white letters spelling “KEEP CALM” rise majestically from the sea’s edge, their clean, modern font contrasting with the organic beauty of nature while reinforcing the tranquil mood, as if the very landscape itself is whispering a soothing mantra of peace, relaxation, and escape from the chaos of everyday life.

    \item \textbf{Sample-10}. A dazzling, multifaceted purple diamond rests regally upon a shimmering bed of iridescent violet sand, its precisely cut facets catching and refracting beams of ethereal light that radiate from behind, casting a luminous glow across the scene and accentuating the gem’s deep amethyst hue with flashes of electric violet and cool silver highlights; the background dissolves into a dreamy, softly diffused gradient of lavender and indigo, enhancing the jewel’s otherworldly brilliance and making it appear almost suspended in a mystical twilight realm, where every angle of its polished surface seems to whisper secrets of rare beauty and enchanted allure, evoking both luxury and fantasy in a single, captivating moment.

    \item \textbf{Sample-11}. In a rain-slicked, neon-drenched cyberpunk cityscape at night, a mysterious hooded figure stands silhouetted against a kaleidoscope of glowing skyscrapers and pulsating billboards, their face obscured by shadow as they hold aloft a luminous rectangular sign that boldly proclaims “UAE” in vibrant, electric-blue neon lettering, casting an otherworldly glow on their gloved hands and the wet pavement below, where reflections of magenta, cyan, and violet lights ripple across the glossy street like liquid electricity, evoking a futuristic vision of the United Arab Emirates as a nexus of technology, mystery, and urban energy under a dark, rain-streaked sky.

    \item \textbf{Sample-12}. A cybernetic warrior stands resolute in the heart of a rain-lashed, neon-soaked metropolis, his face etched with intricate biomechanical tattoos that glow faintly under the pulsating pink and blue lights of towering holographic billboards, while his eyes are hidden behind sleek, futuristic visor goggles radiating a cool violet-blue luminescence that mirrors the city’s electric pulse; clad in a high-collared, armored black jacket accented with glowing orange circuitry along its seams, he exudes an aura of stoic intensity and technological prowess, as blurred silhouettes of passersby dissolve into the background, their forms swallowed by the misty haze and shimmering reflections on wet pavement, immersing him in a world where humanity and machine merge beneath the ceaseless drizzle and chromatic glow of a dystopian urban dreamscape.

    \item \textbf{Sample-13}. As the sun dips below the horizon, casting a warm golden glow across the sky that fades into soft blues and purples, Shanghai’s iconic Oriental Pearl Tower stands tall and radiant, its spherical sections glowing with pink and purple hues that mirror the twilight, anchoring the city’s futuristic skyline against a backdrop of sleek glass skyscrapers and modern high-rises; below, the Huangpu River flows gently, reflecting the fading light and the silhouettes of bridges and riverside trees, while lush green foliage along the embankment frames the scene, adding a touch of nature to the urban grandeur, creating a serene yet dynamic panorama where technological marvels and natural beauty converge in perfect harmony at dusk.

    \item \textbf{Sample-14}. Under a brooding, leaden sky that looms heavy with the promise of storm, a colossal wave rises in furious majesty—its dark, churning body sculpted by unseen winds into a towering, curling crest that crashes forward in a froth of white foam and spray, its deep indigo and slate-gray depths hinting at the ocean’s raw, untamed power; above the tumult, a scattered flock of seabirds soars with outstretched wings, their silhouettes stark against the gloom as they ride the turbulent air currents, embodying both freedom and resilience amid nature’s overwhelming force, while the horizon vanishes beneath the swell, leaving only the primal drama of sea and sky locked in eternal, awe-inspiring conflict.

    \item \textbf{Sample-15}. The image showcases a delectable pepperoni pizza presented on a rustic wooden board, set against a dark, textured background that adds a touch of sophistication. The pizza boasts a golden-brown crust with visible char marks from being cooked in a wood-fired oven, indicating a crispy texture. The cheese, melted and slightly browned in spots, blankets the pizza evenly, with some areas showcasing a rich, gooey appearance. The toppings are predominantly pepperoni slices, arranged in a somewhat circular pattern around the edges, while others lie scattered across the surface in various orientations. Each slice of pepperoni is glossy, indicating a fresh, juicy texture, and they are generously placed, making the pizza look hearty and appetizing. Interspersed among the pepperoni slices are small flecks of herbs, likely basil, adding a burst of green color and freshness to the dish. To the right side of the pizza, two fresh basil leaves are artistically placed, their vibrant green hues contrasting beautifully against the warm tones of the pizza and the wooden board. A few more basil leaves can be seen in the foreground at the bottom left corner, scattered more casually than the ones on the pizza itself. There are also a couple of slices of pepperoni lying outside the pizza, further enhancing the visual appeal of the presentation. The overall composition of the image is balanced, with the pizza centrally located, drawing the viewer's attention immediately. The lighting is subtle yet adequate to highlight the textures and colors of the pizza, making it look inviting and mouth-watering. The slight shadows cast by the pizza and basil leaves add depth to the image, creating a three-dimensional feel.

    \item \textbf{Sample-16}. The image depicts a serene night scene at a lively port town. The sky is filled with a bright starry Milky Way galaxy, casting a soft glow over the entire scene. The town features quaint, charming houses with warm yellow lights emanating from their windows, creating a cozy ambiance. At the forefront, there is a group of people gathered around wooden tables, enjoying their time together. They are engaged in conversation and laughter, with cups of coffee or tea in hand. A golden retriever dog sits by one of the tables, adding to the homely atmosphere. To the right, there is a tall streetlight and a small flower arrangement in a pot, further enhancing the quaint charm of the setting. In the background, a harbor is visible with boats anchored, and the town extends with more houses and shops lining the streets, including a bakery sign.
    
    \item \textbf{Sample-17}. From a high vantage point, the sun rises—or sets—in a blaze of golden-orange light that pierces through a dramatic sky streaked with soft pink, lavender, and deep blue clouds, casting long, ethereal shadows across a vast, snow-blanketed landscape of rolling hills and undulating valleys where a winding road snakes like a ribbon through the serene white expanse; frost-kissed shrubs dot the foreground, their dark branches dusted with snow and catching the warm glow, while the distant horizon fades into a hazy, dreamlike mist, blending earth and sky in a tranquil, almost otherworldly winter tableau that evokes both solitude and sublime beauty beneath the celestial spectacle of dawn or dusk.
    
    \item \textbf{Sample-18}. The image captures the majestic Forbidden City in Beijing, China, bathed in the warm hues of a setting sun. The scene is dominated by several large, traditional Chinese buildings with elegant, ornate roofs painted in vibrant reds and golds. These buildings feature numerous golden dragons and intricate carvings, typical of imperial architecture. The main structure in the center is an imposing palace with multiple eaves and large golden pillars, its entrance flanked by smaller pavilions. The central building's roof is adorned with intricate patterns and two large, pointed gables, adding to its grandeur. In front of the palace, a wide, open courtyard stretches out, paved with smooth, light-colored stones and bordered by white stone balustrades. These balustrades are decorated with sculpted figures and floral designs, providing a stark contrast to the dark stone of the buildings behind them. The courtyard is devoid of people, emphasizing the serene and historical atmosphere of the site. To the left, more buildings can be seen, each with their own distinct architectural features, though slightly obscured due to the architectural layout. The sky above is a soft gradient from pale blue at the horizon to a warm orange near the sun, which casts a gentle glow over the entire scene. A few wispy clouds are scattered across the sky, adding depth and dimension to the panoramic view. In the foreground, there is a series of white, stone railings and steps leading up to the palace, guiding the viewer’s eye towards the impressive structure. The entire area is bathed in the soft, golden light of the sunset, creating a peaceful and timeless quality that highlights the historical significance of this famous landmark.

    \item \textbf{Sample-19}. In this serene, sunset-hued beach scene, a woman stands with her back to the viewer, gazing out at the ocean. She has long brown hair tied loosely behind her head and wears a flowing white sleeveless dress that reaches her ankles. She carries a pair of black flip-flops in her right hand. Her light brown and white dog sits attentively beside her on the sandy shore, their brown and white fur contrasting with the warm, golden tones of the setting sun. The beach is bathed in the soft, orange glow of the setting sun, casting long shadows and highlighting the texture of the sand. In the distance, the gentle waves roll onto the shore, with the sun's reflection shimmering on the water. To the left, a sailboat sails across the calm sea, its silhouette silhouetted against the warm sky. A wooden lifeguard chair with a red life buoy stands near the center-right of the scene, next to a blanket with a floral pattern draped over its legs. The beach is dotted with footprints, and tall grasses and shrubs frame the scene. A couple of seagulls fly low in the orange sky, adding to the tranquil atmosphere. In the background, a cliff rises, partially obscuring the view, and a few more sailboats are visible on the horizon.

    \item \textbf{Sample-20}. An ancient Greek philosopher is talking on a wireless headset.

    \item \textbf{Sample-21}. A serene elven woman with pointed ears and intricate silver face art gazes thoughtfully, clad in a dark green gown with gold trim. She stands in a mystical, moonlit forest where glowing blue mushrooms illuminate the shadowy trees around her.

    \item \textbf{Sample-22}. The image depicts a small, well-lit home office setup in a cozy room with beige carpeting. The primary focus is a compact wooden desk positioned against a pale wall. The desk has a simple, light-colored finish and is supported by two metal legs, which appear to be adjustable for height. On the desk, there is a black keyboard and a laptop computer on the right side, along with a closed, black-framed flat-screen monitor to the left of the laptop. A white mouse and a pair of sunglasses rest on the keyboard. A single table lamp with a black shade stands next to the keyboard, casting a warm light over the workspace. To the left of the lamp, a small stack of books or papers rests on the desk surface. A black rolling chair with height-adjustable arms is stationed directly in front of the desk. The chair's wheels are visible, indicating its portability. The computer monitor is accompanied by a webcam mounted above it on the wall. Below the desk, the floor is partially covered with a light-colored rug that contrasts with the carpeting. Adjacent to the desk, there is a potted plant with lush green leaves placed on a small round table or stand. The room's background features a bookshelf filled with various books, some of which are visible through open shelves. A white cushioned armchair sits to the left of the desk, suggesting a cozy nook for relaxation or additional seating. On the wall behind the desk, near the corner, a rectangular mirror reflects part of the room, adding depth to the space. An overhead lighting fixture casts a soft yellow glow from above, highlighting the desk area while keeping the rest of the room dimly lit. The overall color palette includes neutral tones—beige, white, and shades of brown—creating a calming and functional workspace atmosphere.

    \item \textbf{Sample-23}. In a vibrant, arid desert landscape bathed in warm, golden hues of sunset, a group of three individuals ventures through a rugged, canyon-like terrain. The woman at the center, dressed in a practical olive-green safari outfit with rolled-up sleeves, khaki pants, and a belt bag slung over her shoulder, walks confidently towards the camera. Her dark hair is tied up in a bun, and she has a focused expression on her face as she gazes at the ground. A small, playful fox stands beside her, attentively looking ahead. The woman's right hand holds a stainless steel water bottle, and her left arm is relaxed by her side. On the right, a man wearing a wide-brimmed straw hat, beige shirt, and cargo pants stands observing the surroundings, while his young son, dressed in an orange t-shirt and black shorts, looks back at them with a curious expression. The man and his son are positioned slightly behind the woman, who appears to be leading the way. In the foreground, a cactus plant with a yellow bloom adds to the desert ambiance. The background features towering red rock formations and sparse vegetation, including a few Joshua trees and desert scrub. A large eagle soars high above, its wings spread wide against the backdrop of a sky painted with swirling clouds in shades of orange, pink, and purple. The sand beneath their feet is dotted with footprints, suggesting they have been walking for some time. The entire scene is imbued with a sense of adventure and exploration, set against the timeless beauty of a desert canyon under a dramatic sunset sky.

    \item \textbf{Sample-24}. Please generate a realistic image of the traditional Chinese Hotan Jade pendant. The pendant is a round jade brand, with a full color of turquoise. The jade is warm and delicate, and the surface is highly polished but not excessively reflective, showing the oily texture of real jade. A traditional Jiangnan garden landscape painting is carved in relief on the jade plaque: the upper half of the picture shows a group of Chinese style buildings arranged in a staggered manner, with roofs featuring upturned eaves and horsehead walls, and rich details. The buildings are interspersed with delicate elements such as small bridges, flowing water, weeping willows, and rockeries. The overall composition is complex but not messy, presenting a freehand feeling of traditional Chinese painting style. The lower part of the screen is relatively blank, with only winding rivers flowing from bottom to right, enhancing the spatial hierarchy. The pendant is hung on a gray green Chinese woven rope, with a simple and natural knot, tightly woven from multiple strands of fine thread, with a tough texture. It is decorated with a small red coral bead directly above it. The background of the picture is a light gray cloth in the style of physical photography as a reference, which is overall realistic and realistic. The style is modern high-quality still life photography, with clear composition, soft lighting, and focus on the center of the jade plaque, blurring the background details.

    \item \textbf{Sample-25}. A portrait of profound wisdom and quiet contemplation, this elderly man with a long, flowing white beard and deeply lined face is captured in dramatic chiaroscuro lighting against a dark void, his gaze fixed on something unseen beyond the frame.

\end{itemize}

\paragraph{More ``Image-Text-Image" reconstruction results generated by our method.}

\begin{figure*}[t]
    \centering
    \includegraphics[width=0.8\linewidth]{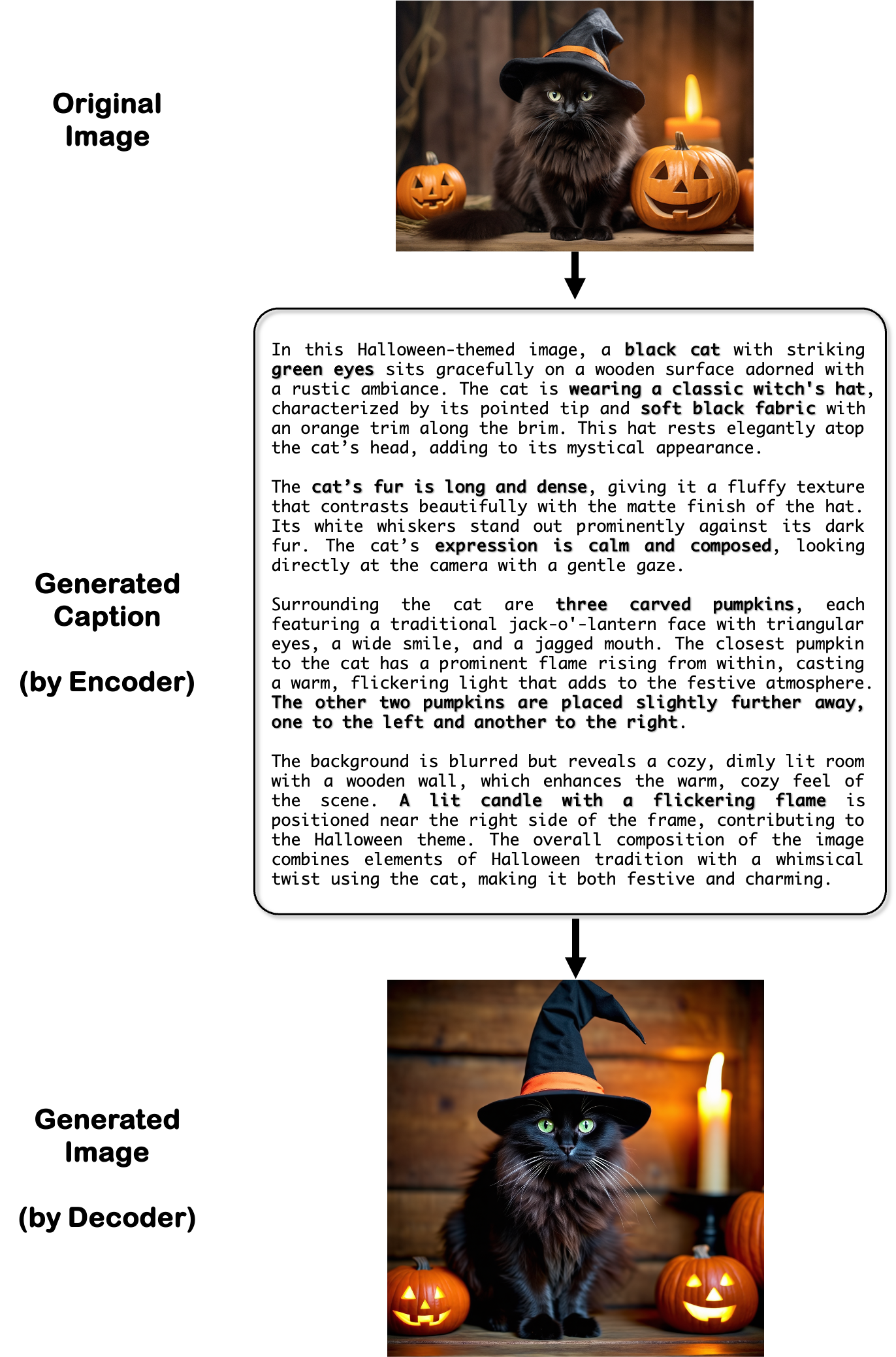}
    \caption{Example of the "image-text-image" reconstruction result by \textbf{our encoder and decoder}.}
    \label{fig:rec_exp_1}
    \vspace{-2mm}
\end{figure*}

\begin{figure*}[t]
    \centering
    \includegraphics[width=0.8\linewidth]{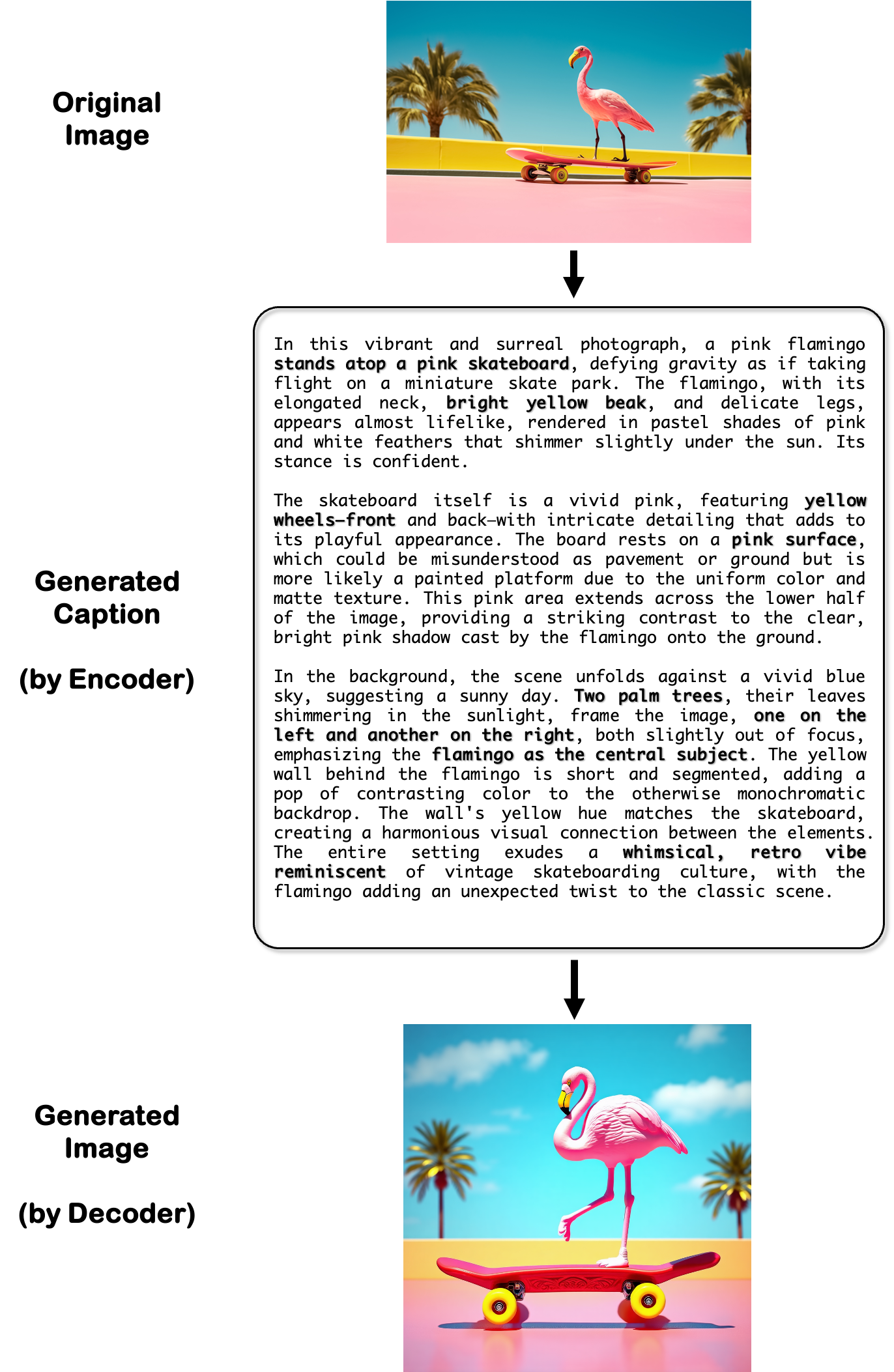}
    \caption{Example of the "image-text-image" reconstruction result by \textbf{our encoder and decoder}.}
    \label{fig:rec_exp_2}
    \vspace{-2mm}
\end{figure*}

\begin{figure*}[t]
    \centering
    \includegraphics[width=0.8\linewidth]{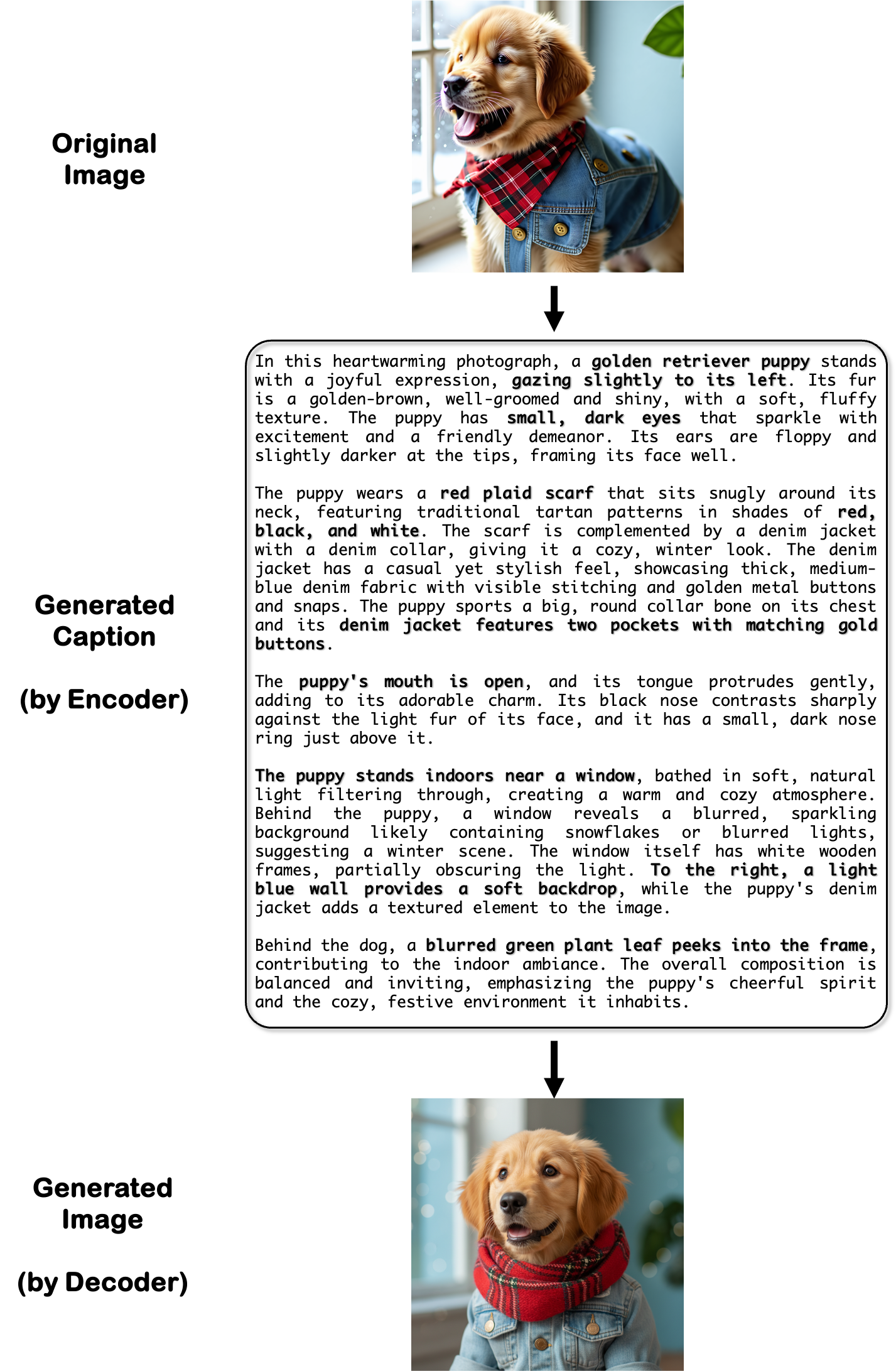}
    \caption{Example of the "image-text-image" reconstruction result by \textbf{our encoder and decoder}.}
    \label{fig:rec_exp_3}
    \vspace{-2mm}
\end{figure*}

Here, we provide more visual examples of the ``image-text-image" reconstruction pipeline using our method, i.e., our encoder processes the input image, generates the output descriptive caption, and then passes it through our decoder to recover it to pixels. See Fig.~\ref{fig:rec_exp_1}, Fig.~\ref{fig:rec_exp_2}, and Fig.~\ref{fig:rec_exp_3} for details.

\paragraph{Scaling unified-bench for testing.}

We have expanded the evaluation by increasing the number of source images to 2000. 
The corresponding results are reported in Tab.~\ref{unifiedbenchv2}, validating the effectiveness of our method.

\begin{table}[h]
    \centering
    \vspace{-3mm}
    \scalebox{0.73}{
    \begin{tabular}{l|c|c|c|c|c}
        \toprule
        Method & Average & CLIP & LongCLIP & DINO-v2 & DINO-v3 \\
        \midrule
        Bagel & 80.95 & 87.57 & 92.88 & 80.11 & 63.24 \\
        GPT-Image* & 83.69 & \textbf{91.37} & 92.81 & \textbf{86.41} & 64.15 \\
        \rowcolor{gray!15} \textbf{Ours} & \textbf{84.74} & 90.22 & \textbf{94.49} & 85.72 & \textbf{68.54} \\
        \bottomrule
    \end{tabular}}
    \vspace{-0.8em}
    \caption{`*' indicates the latest GPT-Image version used for comparison.}
    \vspace{-1.0em}
    \label{unifiedbenchv2}
\end{table}

\paragraph{A deeper analysis of the performance drop on OCR/DU would be beneficial.}

We attribute the performance drop on OCR/DU to \textit{decoder-side text-rendering bottleneck,} rather than to the limitations of our method. 
\begin{wrapfigure}{r}{0.4\linewidth}
    \centering
    \vspace{-1.5em} 
    \scalebox{0.75}{
        \begin{tabular}{l|c|c}
            \toprule
            Decoder  & OCR & DU \\
            \midrule
            \textit{Enc. Only}   & 75.0 & 70.0 \\
            SD3      & 68.8 & 58.4 \\
            Qwen-Image       & 75.9 & 70.5 \\ 
            \bottomrule
        \end{tabular}
    }
    \vspace{-3mm}
    \captionof{table}{Decoder choices for better OCR understanding.} 
    \label{tab:comparison}
    \vspace{-3mm} 
\end{wrapfigure}
In our setup, the decoder is instantiated with \textbf{SD3,} whose limited text generation capability yields a \textbf{noisy supervision signal} for the understanding model, thereby degrading OCR/DU performance.
To verify, we replace SD3 with Qwen-Image (with stronger text rendering). As shown in Tab.~\ref{tab:comparison}, the performance drop \textbf{completely disappears.}

\paragraph{MSE as the reward model.}

We train the unified system under an image-text-image (I2T2I) pipeline and use text as an intermediate representation; \textbf{only semantic content is preserved.} The CLIP model naturally captures this semantic similarity.
Instead, \textbf{MSE operates at the pixel level} and penalizes low-level variations (lighting, texture, style) that do \textit{not affect semantics.}
Given this reason, our main paper utilizes the semantic encoder (CLIP) as the reward model, rather than pixel-level supervision.

\paragraph{The required T2I model/capability at the beginning of training.}

The \textbf{T2I model is expected to possess basic long-instruction following capability,} as our reconstructive RL process progressively produces more detailed intermediate captions. 
If the T2I model or text encoder is constrained by short token limits (e.g., CLIP-style 77-token encoders), it can bottleneck reconstruction quality and thus limit effective RL optimization.
Importantly, this assumption is \textit{not restrictive in practice,} as modern T2I foundations (e.g., Qwen/Longcat-Image) already show strong capacity for handling long and structured textual prompts, and thus naturally satisfy this and make our method \textit{broadly applicable.}

\paragraph{Limitations and future works.}

In our experiment, we observe an unexpected decrease in understanding performance on \textit{visual-text recognition related tasks} (Tab. 7 in the main paper), with accuracy dropping by approximately 10\% on Document Understanding (DU) and OCR. 
This likely stems from the well-known difficulty of current image generation models in faithfully rendering text~\cite{dalle2}, which may introduce misleading reconstruction rewards during RL and hinder the encoder’s ability to learn reliable text semantics. 
This limitation means that the overall benefits of our framework are currently constrained by the weaknesses of the generation module. Improving text reconstruction fidelity is therefore an important direction for future work. 
Our framework naturally extends beyond the image, as it treats text as a universal representation; the same reconstructive principle could be applied to other modalities such as audio and video.

\clearpage

\begin{promptbox}[Prompt for the LLM Sampling]

\textbf{System Prompt:} You are an expert vision-language model. 

\textbf{User Prompt:} Your task is: Given an input image, generate a \textbf{textual description} of the image. If there is text in the image, transcribe it inside double quotes.  \\

Now, carefully analyze the input image and output the full description. \\

Input Image: \{\{image\_path\}\}

\label{llm_prompt}
\end{promptbox}

\begin{promptbox}[Prompt Used to Perform LLM Judge for Caption Quality]

\textbf{User Prompt:}\\
You will conduct a multi-dimensional analysis of each caption based on the specific criteria listed below. For each criterion, you will assign a score from 1 (very poor) to 10 (excellent). After scoring, you must provide a detailed, structured comparative analysis and declare a final winner.

\textbf{Evaluation Criteria \& Scoring:}

Please evaluate each caption against the following four criteria. Provide your scores in a markdown table.

1.  \textbf{Comprehensiveness, Descriptive Richness, and Accuracy:}
\begin{itemize}
    \item How deeply does the caption describe the image? Does it go beyond a superficial glance to include important, specific details (e.g., colors, textures, materials, lighting, background elements, expressions)?
    \item Does it effectively and accurately describe the context (e.g., a black dog not a monkey, or brown eyes not black), environment (background description)?
    \item Does the caption capture subtle nuances that a casual observer might miss?
\end{itemize}

2.  \textbf{Linguistic Fluency and Naturalness:}
\begin{itemize}
    \item Is the caption grammatically correct and well-written in natural-sounding English?
    \item Does it flow like a human would describe the scene, or does it sound robotic, disjointed, or like a list of keywords?
    \item Is the vocabulary choice sophisticated, appropriate, and engaging?
\end{itemize}

3.  \textbf{Semantic and Compositional Insight:}
\begin{itemize}
    \item Does it effectively capture and convey the overall mood, atmosphere, emotion, or narrative implied by the scene?
    \item Does it demonstrate an understanding of the image's composition (e.g., what is in the foreground vs. background)?
\end{itemize}

Based on the above rules, provide a comprehensive, head-to-head comparison of the two captions. Structure your analysis with subheadings for each of the four criteria. For each criterion, explicitly quote phrases from both Caption A and Caption B to illustrate your points and justify the difference in their scores. Explain not just \textit{what} is different, but \textit{why} one caption's approach is superior for describing the provided image.\\

Finally, please declare the winner based on your detailed comparative analysis above. This section must contain only a single letter.\\

\textbf{Final Answer:} [A or B].

\end{promptbox}

\end{document}